\documentclass[11pt]{article}
\usepackage{yub}
\usepackage{bm}
\usepackage[shortlabels]{enumitem}
\usepackage[round, compress]{natbib}

\ifdefined\usebigfont

\usepackage{times}
\usepackage[fontsize=13pt]{scrextend}
\AtBeginDocument{
\usepackage{geometry}	\newgeometry{left=1.56in,right=1.56in,top=1.72in,bottom=1.76in	}
}
\else
\usepackage[margin=1in]{geometry}
\fi

\hypersetup{
    colorlinks,
    linkcolor={blue!50!black},
    citecolor={blue!50!black},
}
\colorlet{linkequation}{blue}

\def\bSigma{{\mathbf \Sigma}}
\def\bA{{\mathbf A}}
\def\bB{{\mathbf B}}
\def\bD{{\mathbf D}}

\def\bW{{\mathbf W}}
\def\bY{{\mathbf Y}}

\def\balpha{{\boldsymbol \alpha}}
\def\bbeta{{\boldsymbol \beta}}
\def\bsigma{{\boldsymbol \sigma}}

\def\ba{{\mathbf a}}

\def\be{{\mathbf e}}

\def\bu{{\mathbf u}}
\def\bv{{\mathbf v}}
\def\bw{{\mathbf w}}
\def\bx{{\mathbf x}}

\def\bzero{{\mathbf 0}}

\def\cD{{\mathcal D}}

\def\cF{{\mathcal F}}

\def\Unif{{\rm Unif}}

\def\opt{{\sf OPT}}

\def\Bx{B_x}
\def\Bw{B_w}
\def\Bwzero{R_{w,0}}
\def\Bwlarge{B_{w,0}}
\def\Bwstar{B_{w,\star}}

\def\balltf{{\sf B}_{2,4}}

\def\Mxop{M_{x,{\rm op}}}

\newcommand{\normtf}[1]{\norm{#1}_{2,4}} 
\newcommand{\normti}[1]{\norm{#1}_{2,\infty}} 
\newcommand{\normttk}[1]{\norm{#1}_{2,2k}} 
\newcommand{\relu}{{\rm relu}}

\ifdefined\usebigfont

\usepackage{times}
\usepackage[fontsize=13pt]{scrextend}
\usepackage[left=1.56in,right=1.56in,top=1.72in,bottom=1.76in]{geometry}
\else
\fi

\def\shownotes{0}  
\ifnum\shownotes=1
\newcommand{\authnote}[2]{$\ll$\textsf{\small #1 notes: #2}$\gg$}
\else
\newcommand{\authnote}[2]{}
\fi

\title{Beyond Linearization: On Quadratic and Higher-Order
  Approximation of Wide Neural Networks}

\author{Yu Bai\thanks{Salesforce
    Research. \texttt{yu.bai@salesforce.com}}
  \and
  Jason D. Lee\thanks{Princeton
    University. \texttt{jasonlee@princeton.edu}}
}

\begin{document}
\maketitle

\newcommand{\sections}{Sections_iclr_revision}
\begin{abstract}
  Recent theoretical work has established connections between
  over-parametrized neural networks and linearized models governed by
  the Neural Tangent Kernels (NTKs). NTK theory leads to concrete
  convergence and generalization results, yet the empirical
  performance of neural networks are observed to exceed their
  linearized models, suggesting insufficiency of this theory.

  Towards closing this gap, we investigate the training of
  over-parametrized neural networks that are beyond the NTK regime yet
  still governed by the Taylor expansion of the network. We bring
  forward the idea of \emph{randomizing} the neural networks, which
  allows them to escape their NTK and couple with quadratic models. We
  show that the optimization landscape of randomized two-layer
  networks are nice and amenable to escaping-saddle algorithms. We
  prove concrete generalization and expressivity results on these
  randomized networks, which leads to sample complexity bounds (of
  learning certain simple functions) that match the NTK and can in
  addition be better by a dimension factor when mild distributional
  assumptions are present. We demonstrate that our randomization
  technique can be generalized systematically beyond the quadratic
  case, by using it to find networks that are coupled with
  higher-order terms in their Taylor series.
  
\end{abstract}

\section{Introduction}


Deep Learning has made remarkable impact on a variety of artificial
intelligence applications such as computer vision, reinforcement
learning, and natural language processing. Though immensely
successful, theoretical understanding of deep learning lags behind. It
is not understood how non-linear neural networks can be efficiently
trained to approximate complex decision boundaries with a relatively
few number of training samples. 

There has been a recent surge of research on connecting neural
networks trained via gradient descent with the neural tangent kernel
(NTK)~\citep{jacot2018neural,du2018gradientb,du2018gradient,chizat2018note,allen2018learning,arora2019exact,arora2019fine}. This
line of analysis proceeds by coupling the training dynamics of the
nonlinear network with the training dynamics of its linearization in a
local neighborhood of the initialization, and then analyzing the
expressiveness and generalization of the network via the corresponding
properties of its linearized model. 

Though powerful, NTK is not yet a completely satisfying theory for
explaining the success of deep learning in practice. In theory, the
expressive power of the linearized model is roughly the same as, and
thus limited to, that of the corresponding random feature
space~\citep{allen2018learning, wei2019regularization} or the
Reproducing Kernel Hilbert Space
(RKHS)~\citep{bietti2019inductive}. While these spaces can approximate
any regular (e.g. bounded Lipschitz) function up to arbitrary
accuracy, the norm of the approximators can be exponentially large in
the feature dimension for certain non-smooth but very simple functions
such as a single ReLU~\citep{yehudai2019power}.  Using NTK analyses,
the sample complexity bound for learning these functions can be poor
whereas experimental evidence suggests that the sample complexity is
mild~\citep{livni2014computational}. In practice, kernel machines with
the NTK have been experimentally demonstrated to yield competitive
results on large-scale tasks such as image classification on CIFAR-10;
yet there is still a non-neglible performance gap between NTK and full
training on the same convolutional
architecture~\citep{arora2019exact,lee2019wide}. It is an increasingly
compelling question whether we can establish theories for training
neural networks beyond the NTK regime.

In this paper, we study the optimization and generalization of
over-parametrized two-layer neural networks via relating to their
\emph{higher-order approximations}, a principled generalization of the
NTK. Our theory starts from the fact that a two-layer neural network
$f_{\bW_0+\bW}(\bx)$ (with smooth activation) can be Taylor expanded
with respect to the weight matrix $\bW$ as
\begin{align*}
  f_{\bW_0+\bW}(\bx) = \sum_{r=1}^m a_r \sigma((\bw_{0,r} +
  \bw_r)^\top\bx) = \underbrace{\sum_{r=1}^m
  a_r\sigma(\bw_{0,r}^\top\bx)}_{f_{\bW_0}(\bx)} +
  \sum_{k=1}^\infty \; \underbrace{\sum_{r=1}^m 
  a_r\frac{\sigma^{(k)}(\bw_{0,r}^\top\bx)}{k!}
  (\bw_r^\top\bx)^k}_{f^{(k)}_{\bW_0,\bW}(\bx)}.
\end{align*}
Above, $f_{\bW_0}$ does not depend on $\bW$, and $f^{(1)}$ corresponds to the NTK model, which is the dominant $\bW$-dependent term when $\set{\bw_r}$ are small and leads to the coupling between the gradient dynamics for training neural net and its NTK $f^{(1)}$.

Our key observation is that the dominance of $f^{(1)}$ is deduced from
comparing the \emph{upper bounds}---rather than the \emph{actual
  values}---of $f^{(k)}_{\bW_0, \bW}(\bx)$. It is a priori possible
that there exists a subset of $\bW$'s in which the dominating term is
not $f^{(1)}$ but some other $f^{(k)}$, $k\ge 2$. If we were able to
train in that set, the gradient dynamics would be coupled with the
dynamics on $f^{(k)}$ rather than $f^{(1)}$ and thus could be very
different. That learning is coupled with $f^{(k)}$ could further offer
possibilities for expressing certain functions with parameters of
lower complexities, or generalizing better, as $f^{(k)}$ is no longer
a linearized model. In this paper, we build on this perspective and
identify concrete regimes in which neural net learning is coupled with
higher-order $f^{(k)}$'s rather than its linearization.


The contribution of this paper can be summarized as follows.
\begin{itemize}[wide, labelwidth=!, labelindent=0pt]
\item We demonstrate that after \emph{randomization}, the linear NTK
  $f^{(1)}$ is no longer the dominant term, and so the gradient
  dynamics of the neural net is no longer coupled with NTK. Through a
  simple sign randomization, the training loss of an over-parametrized
  two-layer neural network can be coupled with that of a quadratic
  model (Section~\ref{section:algorithm}). We prove that the
  randomized neural net loss exhibits a nice optimization landscape in
  that every second-order stationary point has training loss not much
  higher than the best quadratic model, making it amenable to
  efficient minimization (Section~\ref{section:opt}).
\item We establish results on the generalization and expressive power
  of such randomized neural nets (Section~\ref{section:gen}). These
  results lead to sample complexity bounds for learning certain simple
  functions that matches the NTK without distributional assumptions
  and are advantageous when mild isotropic assumptions on the feature
  are present. In particular, using randomized networks, the sample
  complexity bound for learning polynomials (and their linear
  combination) on (relatively) uniform base distributions is $O(d)$
  lower than using NTK.
\item We show that the randomization technique can be generalized to
  find neural nets that are dominated by the $k$-th order term in
  their Taylor series ($k>2$) 
  which we term as \emph{higher-order NTKs}. These models also have
  expressive power similar as the linear NTK, and potentially
  even better generalization and sample complexity
  (Section~\ref{section:higher-order} \&
  Appendix~\ref{appendix:higher-ntk-gen}).
\end{itemize}

\subsection{Prior work}
We review prior work on the optimization, generalization, and expressivity of neural networks.

\paragraph{Neural Net and Kernel Methods}
\cite{neal1996priors} first proposed the connection between
infinite-width networks and kernel methods. Later
work~\citep{daniely2016toward,williams1997computing,
  lee2018deep,novak2019bayesian,matthews2018gaussian} extended this
connection to various settings including deep networks and deep
convolutional networks. These works established that gradient descent
on only the output layer weights is well-approximated by a kernel
method for large width. 

More recently, several groups discovered the connection between
gradient descent on all the parameters and the neural tangent
kernel~\citep{jacot2018neural}. \cite{li2018learning,du2018gradient}
utilized the coupling of the gradient dynamics to prove that gradient
descent finds global minimizers of the training loss of two-layer
networks, and
\cite{du2018gradientb,allen2018convergence,zou2018stochastic}
generalized this to deep residual and convolutional networks. Using
the NTK coupling,
\cite{arora2019fine,cao2019generalizationa,cao2019generalizationb}
proved generalization error bounds that match the kernel method. 

Despite the close theoretical connection between NTK and training deep
networks, \cite{arora2019exact,lee2019wide,chizat2018note} empirically
found a significant performance gap between NTK and actual
training. This gap has been theoretically studied in
\cite{wei2019regularization,allen2019can,yehudai2019power,ghorbani2019limitations}
which established that NTK has provably higher generalization error
than training the neural net for specific data distributions and architectures.

The idea of randomization is initiated by~\citet{allen2018learning},
who use randomization to provably learn a three-layer network; however
it is unclear how the sample complexity of their algorithm compares
against the NTK. Inspired by their work, we study the potential gains
of coupling with a non-linear approximation over the linear NTK --- we
compare the performance of a quadratic approximation model with the
linear NTK on two-layer networks and find that under mild data
assumptions the quadratic approximation reduces sample complexity
under mild data assumptions.



\paragraph{Outside the NTK Regime}
It is believed that the success of SGD is largely due to its
algorithmic regularization effects. A large body of work
\cite{li2017algorithmic,nacson2019lexicographic,gunasekar2018implicit,gunasekar2018characterizing,gunasekar2017implicit,woodworth2019kernel}
shows that asymptotically gradient descent converges to a max-margin
solution with a strong regularization effect, unlike the NTK
regularization\footnote{As a concrete example,
  \cite{woodworth2019kernel} showed that for matrix completion the NTK
  solution estimates zero on all unobserved entries and the max-margin
  solution corresponds to the minimum nuclear norm solution.}. 

 For two-layer networks, a series of works used the mean field method
 to establish the evolution of the network parameters via a
 Wasserstein gradient
 flow~\citep{mei2018mean,chizat2018global,wei2018margin,rotskoff2018neural,sirignano2018mean}.
 In the mean field regime, the parameters move significantly from
 their initialization, unlike NTK regime, however it is unclear if the
 dynamics converge to solutions of low training loss.

Finally, \cite{li2019towards} showed how a combination of large
learning rate and injected noise amplifies the regularization from the
noise and outperforms the NTK of the corresponding architecture. 

\paragraph{Landscape Analysis} Many prior works have tried to
establish favorable landscape properties such as every local minimum
is a global
minimum~\citep{ge2017learning,du2018power,soltanolkotabi2018theoretical,hardt2016identity,freeman2016topology,nguyen2017loss,nguyen2017loss2,
  haeffele2015global, venturi2018neural}. Combining with existing
advances in gradient descent avoiding
saddle-points~\citep{ge2015escaping,lee2016gradient,jin2017escape},
these show that gradient descent find the global minimum. Notably,
\cite{du2018power,ge2017learning} show that gradient descent converges
to solutions also of low test error, with lower sample complexity than
their corresponding NTKs.

\paragraph{Complexity Bounds} Recently, researchers have studied
norm-based generalization
based~\citep{bartlett2017spectrally,neyshabur2015norm,golowich2017size},
tighter compression-based bounds~\citep{arora2018stronger}, and
PAC-Bayes bounds~\citep{dziugaite2017computing,neyshabur2017pac} that
identify properties of the parameter that allow for efficient
generalization. 
\section{Preliminaries}
\label{section:prelim}
\paragraph{Problem setup}
We consider the standard supervised learning task, in which we are
given a labeled dataset $\cD=\set{(\bx_1, y_1), \dots, (\bx_n, y_n)}$,
where $(x_i, y_i)\in\mc{X}\times \mc{Y}$ are sampled i.i.d. from some
distribution $\P$, and we wish to find a predictor $f:\mc{X}\to \mc{Y}$.
Without loss of generality, we assume that
$\mc{X}=\S^{d-1}(\Bx)\subset \R^d$ for some $\Bx>0$ (so that the
features are $d$-dimensional with norm $\Bx$.)

Let $\ell:\mc{Y}\times\R \to\R_{\ge 0}$ be a loss function such that
$\ell(y,0)\le 1$, and $z\mapsto \ell(y, z)$ is convex, 1-Lipschitz,
and three-times differentiable with the second and third derivatives
bounded by one for all $y\in\mc{Y}$.  This includes for example the
logistic and soft hinge loss for classification.
We let
\[
  L(f) \defeq \E_{\cD}[\ell(y, f(\bx))] \defeq \frac{1}{n}\sum_{i=1}^n
  \ell(y_i, f(\bx_i))~~~{\rm and}~~~L_P(f) \defeq \E_{(\bx,
    y)\sim P}[\ell(y, f(\bx))]
\]
denote respectively the empirical risk and population risk for any
predictor $f:\mc{X}\to \mc{Y}$.

\paragraph{Over-parametrized two-layer neural network}
We consider learning an over-parametrized two-layer neural network
of the form
\begin{equation}
  f_\bW(\bx) = f_{\ba,\bW}(\bx) \defeq \frac{1}{\sqrt{m}}
  \ba^\top\sigma(\bW^\top\bx) = \frac{1}{\sqrt{m}}\sum_{r=1}^m
  a_r\sigma(\bw_r^\top \bx), 
\end{equation}
where $\bW=[\bw_1,\dots,\bw_r]\in\R^{d\times m}$ is the first layer
and $\ba=[a_1,\dots,a_m]^\top\in\R^m$ is the second layer.  The
$1/\sqrt{m}$ factor is chosen to account for the effect of
over-parametrization and is consistent with the NTK-type scaling
of~\citep{du2018gradient, arora2019fine}.  In this paper we fix $\ba$
and only train $\bW$ 
(and thus use $f_\bW$ to denote the network.)

Throughout this paper we assume that the activation is second-order
smooth in the following sense.
\begin{assumption}[Smooth activation]
  \label{assumption:weaker-sigma}
  The activation function $\sigma\in C^2(\R)$, and there exists some
  absolute constant $C>0$ such that
  $|\sigma'(t)|\le Ct^2$, $|\sigma''(t)|\le C|t|$, and
  $\sigma''(\cdot)$ is $C$-Lipschitz.
\end{assumption}
An example 
is the cubic ReLU
$\sigma(t)=\relu^3(t)=\max\set{t,0}^3$. The reason for requiring
$\sigma$ to be higher-order smooth (and thus excluding ReLU) will be
made clear in the subsequent text\footnote{We note that the only restrictive
requirement in Assumption~\ref{assumption:weaker-sigma} is the
Lipschitzness of $\sigma''$, which guarantees second-order smoothness
of the objectives.  The bounds on derivatives (and specifically
their bound near zero) are merely for technical convenience and can be
weakened without hurting the results.}.

\subsection{Notation}
We typically reserve lowercases $a,b,\alpha,\beta,\dots$ for scalars,
bold lowercases $\ba,{\mathbf b}, \balpha, \bbeta, \dots$ for vectors,
and bold uppercases $\bA,\bB,\dots$ for matrices. For a matrix
$\bA=[\ba_1,\dots,\ba_m]\in\R^{d\times m}$, its $2,p$ norm is defined
as 
$ 
\norm{\b A}_{2,p} \defeq \paren{ \sum_{r=1}^m \ltwo{\ba_r}^p }^{1/p}
$
for all $p\in[1,\infty]$. In particular we have
$\norm{\cdot}_{2,2}=\lfro{\cdot}$. We let
$\ball_{2,p}(R)\defeq \{\bW:\norm{\bW}_{2,p}\le R\}$ denote a
$2,p$-norm ball of radius $R$. We use standard Big-Oh notation:
$a = O(b)$ for stating $a\le Cb$ for some absolute constant $C>0$, and
$a=\wt{O}(b)$ for $a\le Cb$ where $C$ depends at most logarithmically
in $b$ and all other problem parameters. For a twice-differentiable
function $f:\R^d\to\R$, $\bx_\star$ is called a second-order
stationary point if $\grad f(\bx_\star)=\bzero$ and
$\grad^2 f(\bx_\star)\succeq \bzero$.


\section{Escaping NTK via randomization}
\label{section:algorithm}
To motivate our 
study, we now briefly review the NTK theory for over-parametrized
neural nets and provide insights on how to go beyond the NTK regime.

Let $\bW_0$ denote the weights in a two-layer neural network at
initialization and $\bW$ denote its movement from $\bW_0$ (so that the
current weight matrix is $\bW_0+\bW$.)  The observation in NTK
theory, or the theory of lazy training~\citep{chizat2018note}, is that
for small $\bW$ the neural network $f_{\bW_0+\bW}$ can be Taylor
expanded as
\begin{align*}
  & \quad f_{\bW_0+\bW}(\bx) = \frac{1}{\sqrt{m}}\sum_{r\le m}
    a_r\sigma((\bw_{0,r}+\bw_r)^\top\bx) \\
  & = \underbrace{\frac{1}{\sqrt{m}}\sum_{r\le m}
    a_r\sigma(\bw_{0,r}^\top\bx)}_{f_{\bW_0}(\bx)} +
    \underbrace{\frac{1}{\sqrt{m}}\sum_{r\le m} 
    a_r\sigma'(\bw_{0,r}^\top\bx)(\bw_r^\top\bx)}_{\defeq
    f^L_\bW(\bx)} + O \paren{ \frac{1}{\sqrt{m}}\sum_{r\le m}
    (\bw_r^\top\bx)^2}, 
\end{align*}
so that the network can be decomposed as the sum of the initial
network $f_{\bW_0}$, the linearized model $f^L_{\bW}$, and higher order
terms. Specifically (ignoring $f_{\bW_0}$ for the moment), when $m$ is
large and $\ltwo{\bw_r}=O(m^{-1/2})$, we expect $f^L_{\bW}=O(1)$ and
higher order terms to be $o_m(1)$, which is indeed the regime when we
train $f_{\bW_0+\bW}$ via gradient descent.
Therefore, the trajectory of training $f_{\bW_0+\bW}$ is coupled with the
trajectory of training $f_{\bW_0}+f^L_\bW$, which is a convex problem
and enjoys convergence guarantees~\citep{du2018gradient}.

Our goal is to find subsets of $\bW$ so that the dominating term is
not $f^L$ but something else in the higher order part. The above
expansion makes clear that this cannot be achieved through simple
fixes such as tuning the leading scale $1/\sqrt{m}$ or the learning
rate --- the domination of $f^L$ appears to hold so long as the
movements $\bw_r$ are small.

\paragraph{Randomized coupling with quadratic model}
We now explain how the idea of \emph{randomization},
initiated in~\citep{allen2018learning}, can help get rid of the domination
of $f^L$. 
Let $\bW$ be a fixed weight matrix. Suppose for each weight vector
$\bw_r$, we sample a random variable $\Sigma_{rr}\in\R$ and consider
instead the random weight matrix
\begin{equation*}
  \bW\bSigma \defeq \bW\diag(\set{\Sigma_{rr}}_{r=1}^m) =
  [\Sigma_{11}\bw_1, \dots, \Sigma_{rr}\bw_r],
\end{equation*}
then the second-order Taylor expansion of $f_{\bW_0+\bW\bSigma}$ can
be written as
\begin{align*}
  & \quad f_{\bW_0+\bW\bSigma}(\bx) = \frac{1}{\sqrt{m}}\sum_{r=1}^m
    a_r\sigma\paren{(\bw_{0,r} + \bw_r\Sigma_{rr})^\top\bx} \\
  & = f_{\bW_0}(\bx) +
    \underbrace{\frac{1}{\sqrt{m}}\sum_{r=1}^m
    a_r\sigma'(\bw_{0,r}^\top\bx)(\Sigma_{rr}\bw_r^\top\bx)}_{=
    f^L_{\bW\bSigma}(\bx)} + 
    \underbrace{\frac{1}{2\sqrt{m}}\sum_{r=1}^ma_r\sigma''(\bw_{0,r}^\top\bx)
    \Sigma_{rr}^2(\bw_r^\top\bx)^2}_{\defeq f^Q_{\bW\bSigma}(\bx)} + \dots,
\end{align*}
where we have defined in addition the quadratic part
$f^Q_{\bW\bSigma}$. Due to the existence of $\{\Sigma_{rr}\}$, each
original weight $\bw_r$ now has an additional a scalar that is
different in $f^L$ and $f^Q$. Specifically, if we choose
\begin{equation}
  \label{equation:random-sign}
  \Sigma_{rr} \simiid \Unif\set{\pm 1}
\end{equation}
to be \emph{random signs}, then we have $\Sigma_{rr}^2\equiv 1$ and
thus $f^Q_{\bW\bSigma}(\bx) \equiv f^Q_\bW(\bx)$, whereas
$\E[\Sigma_{rr}]=0$ so that $\E[f^L_{\bW\bSigma}(\bx)]\equiv
0$. Consequently, $f^Q$ is not affected by such randomization whereas
$f^L_{\bW\bSigma}$ is now mean zero and thus can have substantially
lower magnitude than $f^L_\bW$.

More precisely, when $\ltwo{\bw_r}\asymp m^{-1/4}$, the scalings of
$f^L$ and $f^Q$ compare as follows:
\begin{itemize}[wide, labelwidth=!, labelindent=5pt]
\item We have $\E_\bSigma[f^L_{\bW\bSigma}(\bx)]=0$ and
\begin{equation*}
  \E_\bSigma\left[ (f^L_{\bW\bSigma}(\bx))^2 \right] =
  \frac{1}{m}\sum_{r=1}^m
  a_r^2\sigma'(\bw_{0,r}^\top\bx)^2(\bw_r^\top\bx)^2
  = O\left(\frac{1}{m}\sum_{r=1}^m \ltwo{\bw_r}^2\right) =
  O(m^{-1/2}), 
\end{equation*}
so we expect
$
  f^L_{\bW\bSigma}(\bx) = O(m^{-1/4})
$
over a random draw of $\bSigma$.
\item The quadratic part scales as
\begin{equation*}
  f^Q_{\bW\bSigma}(\bx) = f^Q_\bW(\bx) = \frac{1}{2\sqrt{m}}
  \sum_{r=1}^m a_r
  \sigma''(\bw_{0,r}^\top\bx) (\bw_r^\top\bx)^2 = O\left(
    \frac{1}{\sqrt{m}}\sum_{r=1}^m \ltwo{\bw_r}^2 \right) = O(1).
\end{equation*}
\end{itemize}
Therefore, at the random weight matrix $\bW\bSigma$, $f^Q$ dominates
$f^L$ and thus the network is coupled with its quadratic part rather
than the linear NTK. 

\subsection{Learning randomized neural nets}
The randomization technique leads to the following recipe for learning
$\bW$: train $\bW$ so that $\ltwo{\bw_r}= O(m^{-1/4})$ and
$\bW\bSigma$ has in expectation low loss. We make this precise by
formulating the problem as minimizing a \emph{randomized neural net
  risk}.

\paragraph{Randomized risk}
Let $\wt{L}:\R^{d\times m}\to\R$ denote the vanilla empirical risk for
learning $f_\bW$:
\begin{equation*}
  \wt{L}(\bW) = \E_\cD\left[ \ell(y, f_{\bW_0 + \bW}(\bx)) \right],
\end{equation*}
where we have reparametrized the weight matrix into $\bW_0+\bW$ so
that learning starts at $\bW=\bzero$.

Following our randomization recipe, we now formulate our problem as
minimizing the expected risk
\begin{equation*}
  L(\bW) \defeq \E_\bSigma[\wt{L}(\bW\bSigma)] = \E_{\bSigma, 
    \cD}\left[\ell(y, f_{\bW_0 + \bW\bSigma}(\bx))\right],
\end{equation*}
where $\bSigma\in\R^{m\times m}$ is a diagonal matrix with
$\Sigma_{rr}\simiid\Unif\set {\pm 1}$.
To encourage $\ltwo{\bw_r}=O(m^{-1/4})$ and improve generalization, we
consider a regularized version of $\wt{L}$ and $L$ with $\ell_{2,4}$
regularization: 
\begin{equation*}
  \wt{L}_\lambda(\bW) \defeq \wt{L}(\bW) +
  \lambda\normtf{\bW}^8~~~{\rm and}~~~L_\lambda(\bW) =
  L(\bW) + \lambda\normtf{\bW}^8 =
  \E_{\bSigma}[\wt{L}_\lambda(\bW\bSigma)].
\end{equation*}
Our regularizer penalizes $\bW$, i.e. the distance from
initialization, similar as
in~\citep{hu2019understanding}.\footnote{Our specific choice of
  $\normtf{\cdot}$ norm is needed for measuring the
average magnitude of $f^Q_\bW$, whereas the high
(8-th) power is not essential and can be replaced by any $(4+\eps)$-th
power without affecting the result.}

\vspace{-0.2cm}
\paragraph{Symmetric initialization}
We initialize the parameters $(\ba,\bW_0)$ randomly in the following
way: set
\begin{equation}
  \label{equation:symmetric-init}
  \begin{aligned}
    & a_1=\dots=a_{m/2}=+1,~~a_{m/2+1}=\dots=a_m=-1, \\
    & \bw_{0,r}=\bw_{0,r+m/2}\simiid \normal\paren{0,
      \Bx^{-2}I_d},~\forall r\in[m/2].
  \end{aligned}
\end{equation}
Above, we set half of the $a_i$'s as $+1$ and half as $-1$, and the
weights $\bw_{0,r}$ are i.i.d. in the $+1$ half and copied exactly
into the $-1$ half. Such an initialization is almost equivalent to
i.i.d. random $\bW_0$, but has the additional benefit that
$f_{\bW_0}(\bx)\equiv 0$ and also leads to simple expressivity
arguments. Our initialization scale $\Bx^{-2}$ is chosen so that for a
random draw of $\bw_0$, we have $\bw_0^\top\bx\sim\normal(0,1)$, which
is on average $O(1)$\footnote{Our choice covers two commonly used
  scales in neural net analyses: $\Bx=1$,
  $\bw_{0,r}\sim\normal(0, I_d)$ in
  e.g.~\citep{arora2019fine,allen2018learning}; $\Bx=\sqrt{d}$,
  $\bw_{0,r}\sim\normal(0, I_d/d)$ in
  e.g.~\citep{ghorbani2019linearized}.}. For technical convenience, we
also assume henceforth that the realized
$\set{\bw_{0,r}}$ satisfies the bound
\begin{equation}
  \label{equation:w0-bound}
  \max_{r\in[m]}(\Bx\ltwo{\bw_{0,r}}) =
  O\paren{\sqrt{d+\log(m/\delta)}} =  \wt{O}(\sqrt{d}).
\end{equation}
This happens with probability at
least $1-\delta$ under random initialization (see proof in
Appendix~\ref{appendix:w0-bound}), and ensures that
$\max_{r\in[m]}|\bw_{0,r}^\top\bx|\le \wt{O}(\sqrt{d})$ simultaneously
for all $\bx$. 


\section{Optimization}
\label{section:opt}
In this section, we show that $L_\lambda$ enjoys a nice optimization
landscape. 

\subsection{Nice landscape of clean risk}
As the randomized loss $L$ induces coupling of
the neural net $f_{\bW_0+\bW\bSigma}$ with the quadratic model
$f^Q_\bW$, we expect its behavior to resemble the behavior of gradient
descent on the following \emph{clean risk}:
\begin{equation*}
  L^Q(\bW) \defeq \frac{1}{n}\sum_{i=1}^n \ell(y_i, f^Q_\bW(\bx_i)) =
    \frac{1}{n}\sum_{i=1}^n \ell\paren{y_i, 
    \frac{1}{2\sqrt{m}}\<\bx_i\bx_i^\top, \bW
    \bD_i\bW^\top\>}.
\end{equation*}
Above, we have defined diagonal matrices
$\bD_i=\diag(\set{a_r\sigma''(\bw_{0,r}^\top\bx_i)}_{r\in[m]})\in\R^{m\times
  m}$ which are not trained.

We now show that the clean risk $L^Q$, albeit non-convex, possesses a
nice optimization landscape.
\begin{lemma}[Landscape of clean risk]
  \label{lemma:clean-risk}
  Suppose there exists $\bW_\star\in\R^{d\times m}$ such that
  $L^Q(\bW_\star)\le \opt$. Let $\bSigma'\in\R^{m\times m}$ be a
  diagonal matrix with $\Sigma'_{rr}\simiid\Unif\set{\pm 1}$, then we
  have
  \begin{equation}
    \label{equation:good-landscape}
    \begin{aligned}
      & \quad \E_{\bSigma'}\brac{\grad^2 L^Q(\bW)[\bW_\star\bSigma',
        \bW_\star\bSigma']} \\
      & \le \<\grad L^Q(\bW), \bW\> - 2(L^Q(\bW) - \opt) + 
      \wt{O}\paren{d\Bx^4\normtf{\bW}^2\normtf{\bW_\star}^2m^{-1}}.
    \end{aligned}
  \end{equation}
\end{lemma}
This result implies that, for $\bW$ in a certain ball and large $m$,
every point of higher loss than $\bW_\star$ will have either a
first-order or a second-order descent direction. In other words, every
approximate second-order stationary point of $L^Q$ is also an
approximate global minimum.  Our proof utilizes the fact that $L^Q$ is
similar to the loss function in matrix sensing / learning quadratic
neural networks, and builds on recent understandings that the
landscapes of these problems are often
nice~\citep{soltanolkotabi2018theoretical,du2018power,allen2018learning}. The
proof is deferred to Appendix~\ref{appendix:proof-clean-risk}.

\subsection{Nice landscape of randomized neural net risk}
With the coupling between $f_{\bW_0+\bW\bSigma}(\bx)$ and
$f^Q_\bW(\bx)$ in hand, we expect the risk
$L(\bW)=\E[\wt{L}(\bW\bSigma)]$ to enjoy similar guarantees as the
clean risk does $L^Q(\bW)$ in Lemma~\ref{lemma:clean-risk}. 
We make this precise in the following result.
\begin{theorem}[Landscape of $L$]
  \label{theorem:good-landscape-l}
  Suppose there exists $\bW_\star\in\balltf(\Bwstar)$ such that
  $L^Q(\bW_\star)\le \opt$, and that
  \begin{equation}
    \label{equation:good-landscape-m}
    \begin{aligned}
      m \ge O\paren{
      \brac{\Bx^{12}\Bw^{12} + d^4\Bx^4\Bw^4 +
      d^2\Bx^{20}\Bw^{20}}\eps^{-4}
      + d^5\Bx^8\Bw^8\eps^{-2}}.
    \end{aligned}
  \end{equation}
  for some fixed $\eps\in(0,1]$ and $\Bw\ge \Bwstar$,
  then for all $\bW\in\balltf(\Bw)$, we have  
  \begin{equation}
    \label{equation:good-landscape-l}
    \E_{\bSigma'}\brac{\grad^2 L(\bW)[\bW_\star\bSigma',
      \bW_\star\bSigma']} \le \<\grad L(\bW), \bW\> - 2(L(\bW) - \opt)
    + \eps.
  \end{equation}
\end{theorem}
As an immediate corollary, we have a similar characterization of the
regularized loss $L_\lambda$.
\begin{corollary}[Landscape of $L_\lambda$]
  \label{corollary:good-landscape-l-lambda}
  For any $\Bw\ge \Bwstar$, under the conditions of
  Theorem~\ref{theorem:good-landscape-l}, we have for all $\lambda>0$
  and all $\bW\in\balltf(\Bw)$ that
  \begin{equation}
    \begin{aligned}
      & \quad \E_{\bSigma'}\brac{\grad^2
        L_\lambda(\bW)[\bW_\star\bSigma', 
        \bW_\star\bSigma']} \\
      & \le \<\grad L_\lambda(\bW), \bW\> -
      2(L_\lambda(\bW) - \opt) - \lambda\normtf{\bW}^8 +
      C\lambda\normtf{\bW_\star}^8 + \eps,
    \end{aligned}
  \end{equation}
  where $C=O(1)$ is an absolute constant.
\end{corollary}
Theorem~\ref{theorem:good-landscape-l} follows directly from
Lemma~\ref{lemma:clean-risk} through the coupling between $L$ and
$L^Q$ (as well as their gradients and Hessians).
Corollary~\ref{corollary:good-landscape-l-lambda} then follows by
controlling in addition the effect of the regularizer. The full proof of
Theorem~\ref{theorem:good-landscape-l} and
Corollary~\ref{corollary:good-landscape-l-lambda} are deferred to
Appendices~\ref{appendix:proof-good-landscape-l}
and~\ref{appendix:proof-good-landscape-l-lambda}. 

We now present our main optimization result, which follows directly
from Corollary~\ref{corollary:good-landscape-l-lambda}.
\begin{theorem}[Optimization of $L_\lambda$]
  \label{theorem:main}
  Suppose there exists $\bW_\star$ such that
  \begin{equation}
    L^Q(\bW_\star)\le \opt~~~{\rm
      and}~~~\normtf{\bW_\star}\le\Bwstar
  \end{equation}
  for some $\opt>0$. For any $\gamma=\Theta(1)$ and $\eps>0$, we can
  choose $\lambda$ suitably and
  $m\ge \wt{O}({\rm poly}(d, \Bx\Bwstar, \eps^{-1}))$
  such that the regularized loss $L_\lambda$ satisfies the following:
  any second order stationary point $\what{\bW}$ has low loss and
  bounded norm:
  \begin{equation}
    L_\lambda(\what{\bW}) \le (1+\gamma)\opt + \eps~~~{\rm
      and}~~~\normtf{\what{\bW}} \le O(\Bwstar).
  \end{equation}
\end{theorem}
{\bf Proof sketch.}
The proof of Theorem~\ref{theorem:main} consists of two stages: first
``localize'' any second-order stationary point into a (potentially
very big) norm ball using the $\normtf{\cdot}^8$ regularizer, then use
Corollary~\ref{corollary:good-landscape-l-lambda} in this ball to
further deduce that $L_\lambda$ is low and $\normtf{\what{\bW}}\le
O(\normtf{\bW_\star})$. The full proof is deferred to
Appendix~\ref{appendix:proof-main}.

\vspace{-0.2cm}
\paragraph{Efficient optimization \& allowing large learning rate}
Theorem~\ref{theorem:main} states that when the over-parametrization
is enough, any second-order stationary point (SOSP) $\what{\bW}$ of
$L_\lambda$ has loss competitive with $\opt$, the performance of best
quadratic model. Consequently, algorithms that are able to find SOSPs
(escape saddles) such as noisy SGD~\citep{jin2019stochastic} can
efficiently minimize $L_\lambda$ to up to a multiplicative / additive
factor of $\opt$. Further, by sampling a fresh $\bSigma$ at each
iteration and using the stochastic gradient
$\grad_\bW \wt{L}_\lambda(\bW\bSigma)$ (rather than computing the full
$\grad L_\lambda(\bW)$), the noisy SGD iterates can be computed
efficiently.  We note in passing that our coupling results work in any
$\ell_{2,4}$ ball of $O(1)$ size further allows the use of a
\emph{large learning rate}: as soon as the learning rate is bounded by
$O(m^{1/4})$, we would have $\normtf{\bW_t}\le O(1)$ in a constant
number of iterations, and thus our coupling and landscape results
would hold. This is in contrast with the NTK regime which requires the
learning rate to be bounded by $O(1)$~\citep{du2018gradient}.





\section{Generalization and Expressivity}
\label{section:gen}
We now shift attention to studying the generalization and expressivity
of the (randomized) neural net $\what{\bW}$ learned in
Theorem~\ref{theorem:main}.

\subsection{Generalization}
As $\what{\bW}$ is always coupled (through randomization) with the
quadratic model $f^Q_{\what{\bW}}$, we begin by studying the
generalization of the quadratic model.

\paragraph{Generalization of quadratic models} Let
\begin{equation*}
  \cF^Q(\Bw) \defeq \set{\bx\mapsto f^Q_\bW(\bx): \normtf{\bW} \le
    \Bw}
\end{equation*}
denote the class of quadratic models for $\bW$ in a $\ell_{2,4}$
ball. We first present a lemma that relates the Rademacher complexity
of $\cF^Q(\Bw)$ to the expected operator norm of certain feature maps.

\begin{lemma}[Bounding generalization of $f^Q$ via feature operator
  norm]
  \label{lemma:gen-opnorm}
  For any non-negative loss $\ell$ such that $z\mapsto \ell(y, z)$ is
  1-Lipschitz and $\ell(y,0)\le 1$ for all $y\in\mc{Y}$, we have the
  Rademacher complexity bound
  \begin{align*}
      \E_{\bsigma, \bx}\brac{\sup_{\normtf{\bW}\le \Bw}
      \frac{1}{n}\sum_{i=1}^n \sigma_i\ell(y_i, f^Q_\bW(\bx_i))}
     \le \Bw^2\E_{\bsigma,\bx}\brac{\max_{r\in[m]}
      \opnorm{\frac{1}{n}\sum_{i=1}^n\sigma_i\sigma''(\bw_{0,r}^\top\bx_i)
      \bx_i\bx_i^\top}} + \frac{1}{\sqrt{n}},
  \end{align*}
  where $\sigma_i\simiid\Unif\set{\pm 1}$ are Rademacher variables.
\end{lemma}
\paragraph{Operator norm based generalization}
Lemma~\ref{lemma:gen-opnorm} 
suggests a possibility for
the quadratic model to generalize better than the NTK model: the
Rademacher complexity of $\cF^Q(\Bw)$ depends on the ``feature maps''
$\frac{1}{n}\sum_{i=1}^n\sigma_i\sigma''(\bw_{0,r}^\top\bx_i)
\bx_i\bx_i^\top$ through their matrix operator norm.
Compared with the (naive) Frobenius norm based generalization bounds,
the operator norm is never worse and can be better when additional
structure on $\bx$ is present. The proof of
Lemma~\ref{lemma:gen-opnorm} is deferred to
Appendix~\ref{appendix:proof-gen-opnorm}.

We now state our main generalization bound on the (randomized) neural
net loss $L$, which concretizes the above insight.
\begin{theorem}[Generalization of randomized neural net loss]
  \label{theorem:gen-main}
  For any data-dependent $\what{\bW}$ such that
  $\normtf{\what{\bW}}\le\Bw$, we have
  \begin{align*}
    \E_{\bW_0,\cD}\brac{L(\what{\bW}) - L_P(\what{\bW})}
    \le \wt{O}\paren{
      \frac{\Bx^2\Bw^2\Mxop}{\sqrt{n}} + \frac{1}{\sqrt{n}}} +
      \wt{O}\paren{
      \Bx^3\Bw^3m^{-1/4} + d^2\Bx^2\Bw^2m^{-1/2} 
      },
  \end{align*}
  where
  $\Mxop\defeq \paren{\Bx^{-2}
    \E_{\bx}\brac{\opnorm{\frac{1}{n}\sum_{i=1}^n\bx_i\bx_i^\top}}}^{1/2}$
  is the (rescaled) operator norm of the empirical covariance
  matrix. In particular, $\Mxop\le 1$ always holds; if in addition
  $\bv^\top\bx$ is $K\sqrt{\Var(\bv^\top\bx)}$ sub-Gaussian for all
  $\bv\in\S^{d-1}(1)$ and $\kappa(\Cov(\bx)) \le \kappa$, then
  $\Mxop\le \kappa/\sqrt{d}$ whenever $n\ge O(K^4d)$.


\end{theorem}
The generalization bound in
Theorem~\ref{theorem:gen-main} features two desirable
properties:
\begin{enumerate}[(1),wide]
\item For large $m$ (e.g. $m\gtrsim n^4$), the bound scales at most
  logarithmically with the width $m$, therefore allowing learning with
  small samples and extreme over-parametrization;
\item The main term $\wt{O}(\Bx^2\Bw^2\Mxop/\sqrt{n})$ automatically
  adapts to properties of the feature distribution and can lower the
  generalization error than the naive bound by at most $O(1/\sqrt{d})$
  without requiring us to tune any hyperparameter. Concretely, we have
  $\Mxop\le O(1/\sqrt{d})$ when $\bx$ has an isotropic distribution
  such as $\Unif(\S^{d-1}(\Bx))$ or $\Unif\{\pm \Bx/\sqrt{d}\}^d$.
\end{enumerate}
Theorem~\ref{theorem:gen-main} follows directly from
Lemma~\ref{lemma:gen-opnorm} and a matrix concentration Lemma. The
proof is deferred to Appendix~\ref{appendix:proof-gen-main}.


\subsection{Expressivity and Sample Complexity through Quadratic
  Models}
In order to concretize our generalization result, we now study the
expressive power of quadratic models through the concrete example of
learning functions of the form
$\sum_{j\le k}\alpha_j(\bbeta_j^\top\bx)^{p_j}$, i.e. sum of
``one-directional'' polynomials (for consistency and comparability
with~\citep{arora2019fine}.)

\begin{theorem}[Expressing a sum of polynomials through $f^Q$]
  \label{theorem:quad-net-expressivity-sum}
  Suppose $\set{(a_r, \bw_{0,r})}$ are generated according to the
  symmetric initialization~\eqref{equation:symmetric-init} and we use
  $\sigma(t)=\frac{1}{6}\relu^3(t)$ (so that $\sigma''(t)=\relu(t)$.) If
  $f_\star(\bx)=\sum_{j=1}^k \alpha_j(\bbeta_j^\top\bx)^{p_j}$
  achieves training loss $L(f_\star)\le \eps_0$, where
  $p_j-2\in\set{1}\cup\set{2\ell}_{\ell\ge 0}$,. Then so long as the
  width is sufficiently large:
  \begin{equation*}
    m \ge
    \wt{O}\paren{ndk^2\sum_{j=1}^k
      p_j^3\alpha_j^2(\Bx\ltwo{\bbeta_j})^{2p_j}\eps^{-2}},
  \end{equation*}
  we have with probability at least $1-\delta$ (over $\bW_0$) that
  there exists $\bW_\star\in\R^{d\times m}$ such that
  \begin{equation*}
    L^Q(\bW_\star)\le \opt \defeq \eps_0 + \eps~~~{\rm and}~~~
    \normtf{\bW_\star}^4 \le \Bwstar^4 = O\paren{k\sum_{j=1}^k
      p_j^3\alpha_j^2\Bx^{2(p_j-2)}\ltwo{\bbeta}^{2p_j}\delta^{-1}}.
  \end{equation*}
\end{theorem}
The proof of Theorem~\ref{theorem:quad-net-expressivity-sum} is based on
a reduction from expressing degree $p$ polynomials using quadratic
models to expressing degree $p-2$ polynomials using random feature
models.
The proof
can be found in Appendix~
\ref{appendix:proof-quad-net-expressivity-sum}.

\paragraph{Comparison between quadratic and linearized (NTK) models}
We now illustrate our results in Theorem~\ref{theorem:gen-main}
and~\ref{theorem:quad-net-expressivity-sum}
in three concrete examples, in which we
compare the sample complexity bounds of the randomized
(quadratic) network and the linear NTK when $m$ is sufficiently large.


{\bf Learning a single polynomial}.  Suppose
$f_\star(\bx) = \alpha(\bbeta^\top\bx)^p$ satisfies
$L(f_\star)\le \epsilon$, and we wish to find $\what{\bW}$ with
$O(\eps)$ test loss. By
Theorem~\ref{theorem:quad-net-expressivity-sum} we
can choose $\bW_\star$ such that $L^Q(\bW_\star)\le\opt =2 \eps$, and by
Theorem~\ref{theorem:main}
we can find $\what{\bW}$ such that $L(\what{\bW})\le
L_\lambda(\what{\bW}) \le
3\eps$ and $\|\what{\bW}\|_{2,4} = O(\Bwstar)$. Take $B_x=1$, and
assume $\bx$ is sufficiently isotropic so that $\Mxop
=O(\frac{1}{\sqrt{d}})$, the sample complexity
from Theorem \ref{theorem:gen-main} is
$$
n \ge \wt{O}\Big( \frac{B_x^4 B_w^4 \Mxop^2}{\epsilon^2}\Big) =
\wt{O}\Big(\frac{p^3\alpha^2\ltwo{\bbeta}^{2p}}{d\epsilon^2} \Big) \defeq n_Q. 
$$ 

In contrast, the sample complexity for linear
NTK~\citep{arora2019fine,cao2019generalizationa} to reach $\epsilon$
test loss is
$$
n \ge \wt{O}\Big(\frac{p^2 \alpha^2 \|\bbeta\|_2^{2p}}{\epsilon^2}\Big)
\defeq n_L.
$$
We have $n_Q/n_L=\wt{O}(p/d)$, a reduction by a dimension factor unless
$p\asymp d$. We note that the above comparison is simply comparing
upper bounds, since in general the lower bound on the sample
complexity of linear NTK is unknown.

{\bf Learning a noisy $2$-XOR}.~\citet{wei2019regularization}
established a sample complexity lower bound of linear NTK of
$n \ge n_L = \Omega(d^2)$ to achieve constant generalization error on
the noisy $2$-XOR problem, which allows for a rigorous comparison
against the quadratic model.  

The ground truth function in $2$-XOR is
$f_\star (\bx) = x_1x_2 = ([(\be_1 + \be_2)^\top\bx]^2 - [(\be_1
-\be_2)^\top\bx]^2)/4$, where $\bx\in \{\pm 1\}^d$, and $f_\star$
attains constant margin on the training distribution constructed in
\cite{wei2019regularization}. By
Theorem~\ref{theorem:quad-net-expressivity-sum},
$f_\star$ can be $\eps$-approximated by $f^Q_{\bW_\star}$ with
$\Bwstar^4\le O(1)$. Thus by
Theorem~\ref{theorem:gen-main} the sample complexity for
learning noisy $2$-XOR through the randomized net $\what{\bW}$ is
$$
n \ge n_Q = \wt{O}\paren{\frac{\Bx^4\Bwstar^4\Mxop^2}{\eps^2}} =
\wt{O}\Big(\frac{d}{\epsilon^2}\Big).
$$
This is $\wt{O}(d)$ better than the \textit{sample complexity
  lower bound} of linear NTK and thus provably better.

{\bf Low-rank matrix sensing}. Suppose we wish to learn a symmetric
low-rank matrix $\bA_\star\in\R^{d\times d}$ with
$\opnorm{\bA_\star}\le 1$ and ${\rm rank}(\bA_\star)\le r$
through $n$ rank-one observations of the form
$y_i=f_\star(\bx_i)=\<\bA_\star,\bx_i\bx_i^\top\>$ where $\bx\sim
\Unif(\S^{d-1}(\sqrt{d}))$. This ground truth function can be
written as $f_\star(\bx_i)=\sum_{j=1}^r
\alpha_j(\bv_j^\top\bx_i)^2$ where $|\alpha_j|\le 1$ are the
eigenvalues of $\bA$ and $\bv_j\in\R^d$ are the corresponding
eigenvectors.
For any 1-Lipschitz loss such as the absolute loss, by
Theorem~\ref{theorem:quad-net-expressivity-sum}, there exists
$\bW_\star$ such that $L^Q(\bW_\star)\le \opt = \eps$ and
$\Bwstar^4\le O(r\sum_{j=1}^r \ltwo{\bv_j}^4) = O(r^2)$. Thus by
Theorem~\ref{theorem:gen-main}, 
the sample complexity of reaching $2\eps$ test loss
\begin{equation*}
  \E_\bx\brac{\abs{\<\bA_\star,\bx\bx^\top\> - f_{\what{\bW}}(\bx)}}
  \le 2\eps
\end{equation*}
through the randomized net $\what{\bW}$ is
\begin{equation*}
  n \ge n_Q = \wt{O}\paren{\frac{\Bx^4\Bwstar^4\Mxop^2}{\eps^2}} =
  \wt{O}\paren{\frac{dr^2}{\eps^2}}.
\end{equation*}
This compares favorably against the sample complexity upper bound for
linear NTK, which needs
\begin{equation*}
  n \ge n_L = \wt{O}\paren{\frac{\Bx^4\cdot (\sum_{j=1}^r
      \alpha_j\ltwo{\bv_j}^2)^2}{\eps^2}} =
  \wt{O}\paren{\frac{d^2r^2}{\eps^2}}
\end{equation*}
samples.

\section{Higher-order NTKs}
\label{section:higher-order}
In this section, we demonstrate that our idea of randomization for
changing the dynamics of learning neural networks can be generalized
systematically --- through randomization we are able to obtain
over-parametrized neural networks in which the $k$-th order term
dominates the Taylor series.  Consider a two-layer neural network with
$2m$ neurons and symmetric initialization
(cf.~\eqref{equation:symmetric-init})
\begin{equation*}
  f_{\bW_0+\bW}(\bx) = \frac{1}{\sqrt{m}}\sum_{r\le m}
  \sigma((\bw_{0,r}+\bw_{+,r})^\top\bx) -
  \sigma((\bw_{0,r}+\bw_{-,r})^\top\bx).
\end{equation*}
Assuming $\sigma$ is analytic on $\R$ (i.e. it equals its Taylor
series at any point), we have
\begin{equation*}
  f_{\bW_0+\bW}(\bx) = \sum_{k=0}^\infty f^{(k)}_{\bW_0,\bW}(\bx),
\end{equation*}
where we have defined the \emph{$k$-th order NTK}
\begin{equation*}
  f^{(k)}_{\bW_0, \bW}(\bx) \defeq \frac{1}{\sqrt{m}}\sum_{r\le m}
  \frac{1}{k!} \sigma^{(k)}(\bw_{0,r}^\top\bx)
  \paren{(\bw_{+,r}^\top\bx)^k - (\bw_{-,r}^\top\bx)^k}.
\end{equation*}
Note that $f^{(0)}(\bx)\equiv 0$ due to the symmetric initialization,
and $f^{(1)}(\bx)$ is the standard NTK. For an arbitrary $\bW$ such that
$\ltwo{\bw_{+,r}},\ltwo{\bw_{-,r}}=o_m(1)$, we expect that
$f^{(1)}(\bx)$ is the dominating term in the expansion.

\subsection{Extracting the $k$-th order term}
We now describe an approach to finding $\bW$ so that
\begin{equation*}
  f_{\bW_0+\bW}(\bx) = f^{(k)}_{\bW_0, \bW}(\bx) + o_m(1),
\end{equation*}
that is, the neural net is approximately the $k$-th order NTK plus an
error term that goes to zero as $m\to\infty$, thereby ``escaping'' the
NTK regime. Our approach builds on the following randomization
technique: let $z_+$, $z_-$ be two random variables (distributions)
such that
\begin{equation*}
  \E[z_+^j] = \E[z_-^j]~~{\rm for}~j=0,1,\dots,k-1~~~{\rm
    and}~~~\E[z_+^k] = \E[z_-^k] + 1.
\end{equation*}
Set $(\bw_{+,r},\bw_{-,r}) = (z_{+,r}\bw_{\star,r},
z_{-,r}\bw_{\star,r})$, and take $\ltwo{\bw_{\star,r}}=O(m^{-1/2k})$, we have
\begin{align*}
  f^{(j)}_{\bW_0,\bW}(\bx) = \frac{1}{\sqrt{m}} \sum_{r\le
  m}\frac{1}{j!}\sigma^{(j)}(\bw_{0,r}^\top\bx) \underbrace{(z_{+,r}^j -
  z_{-,r}^j)}_{\textrm{mean
  zero}}\underbrace{(\bw_{\star,r}^\top\bx)^j}_{O(m^{-j/2k})} = O_p(m^{-j/2k})
\end{align*}
for all $j=1,\dots,k-1$, and
\begin{align*}
  & \quad f^{(k)}_{\bW_0,\bW}(\bx) = \frac{1}{\sqrt{m}} \sum_{r\le
    m}\frac{1}{k!}\sigma^{(k)}(\bw_{0,r}^\top\bx) \underbrace{(z_{+,r}^k -
    z_{-,r}^k)}_{\textrm{mean}=1}\underbrace{(\bw_{\star,r}^\top\bx)^k}_{O(m^{-1/2})}
      = O_P(1),
\end{align*}
and
\begin{align*}
  f^{(k+1)}_{\bW_0,\bW}(\bx) = \frac{1}{\sqrt{m}} \sum_{r\le
  m}\frac{1}{(k+1)!}\sigma^{(k+1)}(\bw_{0,r}^\top\bx) \underbrace{(z_{+,r}^{k+1} -
  z_{-,r}^{k+1})}\underbrace{(\bw_{\star,r}^\top\bx)^{k+1}}_{O(m^{-(k+1)/2k})}
  = O_P(m^{-1/2k}).
\end{align*}
Therefore, with high probability, all $f^{(1)},\dots,f^{(k-1)}$ as
well as the remainder term $f-\sum_{j\le k}f^{(j)}$ has order
$O(m^{-1/2k})$, and the $k$-th order NTK $f^{(k)}$ can express an
$O(1)$ function.

\vspace{-0.2cm}

\paragraph{Generalization, expressivity, and ``deterministic'' coupling}
We establish the generalization of expressivity of $f^{(k)}$ in
Appendix~\ref{appendix:higher-ntk-gen}, which systematically extends
our results on the quadratic model. We show that the sample complexity
for learning degree $\ge k$ polynomials through $f^{(k)}$ compared
with linear NTK can be better by a factor of $d^{k-1}$ for large $n$,
when mild distributional assumptions on $\bx$ such as approximate isotropy (constant condition number of the $k^{th}$ moment tensor) is present.

Further, one can extend the concentration arguments on the above randomization to show the existence of some deterministic) $\bW$ at which the neural net is approximately the $k$-th order NTK: $\E_{\bx}[|f_{\bW_0+\bW}(\bx) - f^{(k)}_{\bW_0,\bW}(\bx)|]\le \eps_m$,
where $\eps_m\to 0$ as $m\to\infty$. We would like to leave this as future work.




\section{Conclusion}
In this paper we proposed and studied the optimization and
generalization of over-parametrized neural networks through coupling
with higher-order terms in their Taylor series. Through coupling with
the quadratic model, we showed that the randomized two-layer neural
net has a nice optimization landscape (every second-order stationary
point has low loss) and is thus amenable to efficient minimization
through escape-saddle style algorithms. These networks enjoy the same
expressivity and generalization guarantees as linearized models but in
addition can generalize better by a dimension factor when
distributional assumptions are present. We extended the idea of
randomization to show the existence of neural networks whose Taylor
series is dominated by the $k$-th order term.

We believe our work brings in a number of open questions, such as how
to better utilize the expressivity of quadratic models, or whether the
study of higher-order expansions can lead to a more satisfying theory
for explaining the success of full training. We also note that the
Taylor series is only one avenue to obtaining accurate approximations
of nonlinear neural networks. It would be of interest to design other
approximation schemes for neural networks that are coupled with the
network in larger regions of the parameter space.

\section*{Acknowledgment}
The authors would like to thank Wei Hu, Tengyu Ma, Song Mei, and
Andrea Montanari for their insightful comments.  JDL acknowledges
support of the ARO under MURI Award W911NF-11-1-0303, the Sloan
Research Fellowship, and NSF CCF \#1900145. The majority of this work
was done while YB was at Stanford University.  The authors also thank
the Simons Institute Summer 2019 program on the Foundations of Deep
Learning, and the Institute of Advanced Studies Special Year on
Optimization, Statistics, and Theoretical Machine Learning for hosting
the authors.

\bibliography{bib,addref}
\bibliographystyle{plainnat}

\appendix

\section{Technical tools}
\subsection{A matrix operator norm concentration bound}
\begin{lemma}[Variant of Theroem 4.6.1,~\citep{tropp2015introduction}]
  \label{lemma:matrix-concentration}
  Suppose $\set{\bA_{r,i}}_{r\in[m],i\in[n]}$ are fixed
  symmetric $d\times d$ matrices, and $\set{\sigma_i}_{i\in[n]}\simiid 
  \Unif\set{\pm 1}$ are Rademacher variables. Letting
  \begin{equation*}
    \bY_r = \sum_{i=1}^n \sigma_i\bA_{r,i},
  \end{equation*}
  then we have
  \begin{equation*}
    \E_{\bsigma}\brac{\max_{r\in[m]} \opnorm{\bY_r}}
    \le 4\sqrt{\max_{r\in [m]}v(\bY_r)\log(2md)},
  \end{equation*}
  where
  \begin{equation*}
    v(\bY) \defeq \opnorm{\E_{\bsigma}[\bY^2]}.
  \end{equation*}
\end{lemma}
\begin{proof}
  Applying the high-probability bound in~\citep[Theorem
  4.6.1]{tropp2015introduction} and the union bound, we get
  \begin{align*}
    & \quad \P\paren{\max_{r\in[m]} \opnorm{\bY_r} \ge t} \le 2d\sum_{r\in
      [m]} \exp(-t^2/2v(\bY_r)) \\
    & \le 2dm\exp(-t^2/2\max_{r\in[m]}v(\bY_r)) =
    \exp\paren{-\frac{t^2}{2\max_{r\in[m]}v(\bY_r)} + \log(2dm)}.
  \end{align*}
  Let $V\defeq \max_{r\in[m]}v(\bY_r)$, we have by integrating the
  above bound over $t$ that 
  \begin{align*}
    & \quad \E\brac{\max_{r\in[m]} \opnorm{\bY_r}} \le \int_0^\infty 
      \min\set{\exp\paren{-\frac{t^2}{2V} +
      \log(2dm)}, 1} dt \\
    & \le \sqrt{4V\log(2dm)} + \int_{\sqrt{4V\log(2dm)}}^\infty
      \exp(-t^2/2V + \log(2dm))dt \\
    & \le \sqrt{4V\log(2dm)} + \int_{\sqrt{4V\log(2dm)}}^\infty
      \exp(-t^2/4V)dt \\
    & \le \sqrt{4V\log(2dm)} + \frac{\sqrt{4\pi V}}{2dm} \le
      4\sqrt{V\log(2dm)}. 
  \end{align*}
\end{proof}

\subsection{Expressing polynomials with random features}
\begin{lemma}
  \label{lemma:relu-rf-expressivity}
  Let $\sigma(t)=\relu(t)$ and $\bw_0\sim\normal(0, \Bx^{-2}I_d)$
  be Gaussian random features. For any
  $p\in\set{1}\cup\set{2\ell}_{\ell\ge 0}$ and $\bbeta\in\R^d$, there
  exists a random variable $a=a(\bw_0)$ such that
  \begin{equation*}
    \E_{\bw_0}[\sigma(\bw_0^\top\bx)a] = \alpha(\bbeta^\top\bx)^p
  \end{equation*}
  and $a$ satisfies the $\ell_2$ norm bound
  \begin{equation*}
    \E_{\bw_0}[a^2] \le 2\pi (p\vee
    1)^3\alpha^2 \Bx^{2(p-1)}d\ltwo{\bbeta}^{2p}.
  \end{equation*}
\end{lemma}
\begin{proof}
  Consider the ReLU random feature kernel
  \begin{equation*}
    K(\bx, \bx') =
    \E_{\bw_0\sim\normal(0,
      \Bx^{-2}I_d)}[\relu(\bw_0^\top\bx)\relu(\bw_0^\top\bx')],
  \end{equation*}
  and let $\mc{H}_K$ denote the RKHS associated with this kernel.
  By the equivalence of feature maps~\citep[Proposition
  1]{minh2006mercer}, for any feature map $\phi: \S^{d-1}(\Bx)\mapsto
  \mc{H}$ (where $\mc{H}$ is a Hilbert space) that generates
  $K$ in the sense that
  \begin{equation*}
    K(\bx, \bx') = \<\phi(\bx), \phi(\bx')\>_{\mc{H}},
  \end{equation*}
  we have for any function $f$ that
  \begin{equation}
    \label{equation:feature-map-equivalence}
    \norm{f}_{\mc{H}_K}^2 = \inf_{a\in\mc{H}}
    \set{\norm{a}_{\mc{H}}^2 : f_\star(x) \equiv \<a, \phi(\bx)\>},
  \end{equation}
  and the infimum over $a$ is attainable whenever it is finite.

  For the ReLU random feature kernel $K$, let $u\defeq
  \bx^\top\bx'/\Bx^2$ and $\normal_2(\rho)$ denote a bivariate normal 
  distribution with marginals $\normal(0,1)$ and correlation
  $\rho\in[-1,1]$. We have that 
  \begin{align*}
    & \quad K(\bx, \bx') = \E_{\bw_0\sim\normal(0, \Bx^{-2}I_d)}
      \brac{\relu(\bw_0^\top\bx) \relu(\bw_0^\top\bx')} \\
    & = \E_{(\Z_1, Z_2)\sim\normal_2(u)} [
      \relu(Z_1)\relu(Z_2) ] \\
    & = \frac{1}{2\pi}\paren{u(\pi - \arccos u)
      + \sqrt{1 - u^2}} \\
    & = \frac{1}{2\pi}
      \paren{1 + \frac{\pi}{2}u +
      \sum_{\ell=1}^\infty 
      \frac{(2\ell-3)!!}{(2\ell-2)!!(2\ell-1)(2\ell)} u^{2\ell}} \\
    & = \sum_{p\in\set{0,1}\cup \set{2\ell}_{\ell\ge
      1}} c_p (\bx^\top\bx')^p \Bx^{-2p} \\
    & = \sum_{p\in\set{0,1}\cup \set{2\ell}_{\ell\ge 1}}
      \<\sqrt{c_p}\Bx^{-p}\bx^{\otimes p},
      \sqrt{c_p}\Bx^{-p}(\bx')^{\otimes p} \>,
  \end{align*}
  where the constants $\set{c_p}$ satisfy
  \begin{equation*}
    c_0 = 1/(2\pi), ~~~c_1 = 1/4, ~~~c_{2\ell} \ge
    \frac{1}{2\pi(2\ell-1)^2(2\ell)}~~{\rm for}~\ell \ge 1,
  \end{equation*}
  (and thus $c_p\ge (2\pi (p\vee 1)^3)^{-1}$ for all $p$), 
  and $\bx^{\otimes k}\in\R^{d^k}$ denote the $k$-wise tensor product
  of $\bx$. Therefore, if we define feature map
  \begin{equation*}
    \phi(\bx) \defeq \brac{ \sqrt{c_p}\Bx^{-p}\bx^{\otimes
        p} }_{p\in\set{0,1}\cup\set{2\ell}_{\ell\ge 1}},
  \end{equation*}
  we have $K(\bx,\bx')=\<\phi(\bx), \phi(\bx')\>$. With this feature
  map, the function $f_\star(\bx) = \alpha(\beta^\top\bx)^{p}$ can be
  represented as
  \begin{equation*}
    f_\star(\bx) \equiv \<c_\star, \phi(\bx)\>~~~{\rm
      where}~~~c_\star = \brac{0, \dots, 0, \alpha \cdot
      c_p^{-1/2}\Bx^p \bbeta^{\otimes p}, 0, \dots}.
  \end{equation*}
  Thus by the feature map
  equivalence~\eqref{equation:feature-map-equivalence}, we have
  $f_\star\in\mc{H}_K$ and 
  \begin{equation*}
    \norm{f}_{\mc{H}_K}^2 \le \norm{c_\star}^2 = \alpha^2
    c_p^{-1}\Bx^{2p} \ltwo{\bbeta}^{2p} \le 2\pi (p\vee
    1)^3\alpha^2 \Bx^{2p}\ltwo{\bbeta}^{2p}.
  \end{equation*}
  Now apply the feature map
  equivalence~\eqref{equation:feature-map-equivalence} again with the
  random feature map
  \begin{equation*}
    \bx \mapsto \set{\relu(\bw_0^\top\bx)}_{\bw_0}
  \end{equation*}
  (which maps into the inner product space of $\bw_0\sim\normal(0,
  \Bx^{-2}I_d)$), we conclude that there exists $a=a(\bw_0)$ such that
  $f_\star=\E_{\bw_0}[\relu(\bw_0^\top\bx)a]$ and
  \begin{equation*}
    \E_{\bw_0}[a^2] \le \norm{f_\star}_{\mc{H}_K}^2 \le 2\pi (p\vee
    1)^3\alpha^2 \Bx^{2p}\ltwo{\bbeta}^{2p}.
  \end{equation*}
\end{proof}

\subsection{Proof of Equation~\eqref{equation:w0-bound}}
\label{appendix:w0-bound}
Let $\mc{N}$ be an $1/2$-covering of $\S^{d-1}(1)$. We have
$|\mc{N}|\le 5^d$ and for any vector $\bw\in\R^d$ that
$\ltwo{\bw}\le 2\sup_{\bv\in\mc{N}}(\bv^\top\bw)$ (see
e.g.~\citep[Section A]{mei2018landscape}.) We thus have
\begin{align*}
  & \quad \P\paren{\max_{r\in[m]}\Bx\ltwo{\bw_{0,r}}\ge t} \\
  & \le
    \paren{\max_{\bv\in\mc{N}} \Bx(\bv^\top\bw_{0,r}) \ge t/2} \le 
    \exp(-t^2/8 + \log|\mc{N}| + \log m) \le \exp(-t^2/8+d\log 5 + \log
    m). 
\end{align*}
Setting $t=\sqrt{8(d\log 5+\log
  (m/\delta))}=O(\sqrt{d+\log(m/\delta)})$ ensures that the above
probability does not exceed $\delta$ as desired. \qed
\section{Proofs for Section~\ref{section:opt}}
\subsection{Proof of Lemma~\ref{lemma:clean-risk}}
\label{appendix:proof-clean-risk}


Computing the gradient of $L^Q$, we obtain
\begin{equation*}
  \grad L^Q(\bW) = \frac{2}{n}\sum_{i=1}^n \ell'(y_i, f^Q_\bW(\bx_i))
  \frac{1}{2\sqrt{m}}\bx_i\bx_i^\top\bW\bD_i.
\end{equation*}
Further computing the Hessian gives
\begin{align*}
  \grad^2 L^Q(\bW)[\bW_\star\bSigma', \bW_\star\bSigma']
  & = \frac{2}{n}\sum_{i=1}^n \ell'(y_i, f^Q_\bW(\bx_i)) \cdot
    \underbrace{\frac{1}{2\sqrt{m}}\<\bx_i\bx_i^\top,
    \bW_\star\bSigma'\bD_i\bSigma'\bW_\star^\top\>}_{f^Q_{\bW_\star}(\bx_i)}
  \\ & \qquad+ 
       \frac{4}{n}\sum_{i=1}^n \ell''(y_i, f^Q_\bW(\bx_i)) \cdot
       \Bigg( \underbrace{\frac{1}{2\sqrt{m}}
       \<\bx_i\bx_i^\top,
       \bW\bD_i\bW_\star^\top\bSigma'\>}_{\defeq \wt{y}_i}\Bigg)^2 \\
  & = \underbrace{\frac{2}{n}\sum_{i=1}^n \ell'(y_i,
    f^Q_\bW(\bx_i))f^Q_{\bW_\star}(\bx_i)}_{\rm I} +
    \underbrace{\frac{4}{n}\sum_{i=1}^n \ell''(y_i, f^Q_\bW(\bx_i))
    \wt{y}_i^2}_{\rm II}.  
\end{align*}

Taking expectation over $\bSigma'$, and using that $\ell''\le 1$,
term II can be bounded as 
\begin{align*}
  & \quad \E_{\bSigma'}[{\rm II}] \le
    \E_{\bSigma'}\brac{\frac{4}{n}\sum_{i=1}^n \wt{y}_i^2} \\
  & = \E_{\bSigma',
    \cD} \brac{\frac{2}{m}\sum_{r\le m} 
    \sigma''(\bw_{0,r}^\top\bx)^2(\bw_r^\top\bx)^2
    (\Sigma'_{rr}\bw_{\star,r}^\top\bx)^2} \\
  & \le C\cdot \E_{\cD}\brac{\frac{1}{m}\sum_{r\le m}
    (\bw_{0,r}^\top\bx)^2(\bw_r^\top\bx)^2(\bw_{\star,r}^\top\bx)^2}
  \\
  & \le C \Bx^4\max_{r\in[m], i\in[n]} (\bw_{0,r}^\top\bx_i)^2
    \cdot  \frac{1}{m}\sum_{r\le m}\ltwo{\bw_r}^2\ltwo{\bw_{\star,r}}^2 \\
  &  \le \wt{O}\paren{d\Bx^4\normtf{\bW}^2
    \normtf{\bW_\star}^2m^{-1}}, 
\end{align*}
where the last step used Cauchy-Schwarz on $\set{\ltwo{\bw_r}}$ and
$\set{\ltwo{\bw_{\star,r}}}$.

Term I does not involve $\bSigma'$ and can be deterministically
bounded as
\begin{align*}
  & \quad {\rm I} = 2\E_{\cD}[\ell'(y, f^Q_\bW(\bx))
    f^Q_{\bW_\star}(\bx)] \\
  & = 2\E_{\cD}[\ell'(y, f^Q_\bW(\bx))
    f^Q_\bW(\bx)] + 2\E_{\cD}[\ell'(y, f^Q_\bW(\bx))
    (f^Q_{\bW_\star}(\bx) - f^Q_{\bW}(\bx))] \\
  & \stackrel{(i)}{\le } \<\grad L^Q(\bW), \bW\> + 2\E_{\cD}[\ell(y,
    f^Q_{\bW_\star}(\bx)) -
    \ell(y, f^Q_\bW(\bx))] \\
  & \stackrel{(ii)}{=} \<\grad L^Q(\bW), \bW\> - 2(L^Q(\bW) - \opt). 
\end{align*}
where (i) follows directly by computing $\<\grad L^Q(\bW), \bW\>$ and
the convexity of $z\mapsto \ell(y,z)$, and 
(ii) follows from the assumption that $L^Q(\bW_\star)\le
\opt$.
Combining the bounds for terms I and II gives the desired result. \qed





\subsection{Coupling lemmas}
\begin{lemma}[Bound on $f^Q$]
  \label{lemma:bound-fq}
  For any $\bW\in\R^{d\times m}$, the quadratic model $f^Q_\bW$
  satisfies the bound
  \begin{equation*}
    \abs{f^Q_\bW(\bx)} \le \wt{O}\paren{\sqrt{d}\Bx^2\normtf{\bW}^2}
  \end{equation*}
  for all $\bx\in S^{d-1}(\Bx)$.
\end{lemma}
\begin{proof}
  We have
  \begin{align*}
    & \quad \abs{f^Q_\bW(\bx)} = \abs{\frac{1}{\sqrt{m}} \sum_{r\le m}
      a_r\sigma''(\bw_{0,r}^\top\bx) (\bw_r^\top\bx)^2} \\
    & \le \frac{1}{\sqrt{m}} \sum_{r\le m}
      C\abs{\bw_{0,r}^\top\bx} \cdot (\bw_r^\top\bx)^2
      \le C\sqrt{m}\Bx^2\max_{r\in[m]}\abs{\bw_{0,r}^\top\bx}
      \cdot \frac{1}{m}\sum_{r\le m}\ltwo{\bw_r}^2 \\
    & \le C\sqrt{m}\Bx^2 \wt{O}(\sqrt{d}) \cdot \paren{\frac{1}{m}
      \sum_{r\le m}\ltwo{\bw_r}^4}^{1/2} =
      \wt{O}\paren{\sqrt{d}\Bx^2\Bwzero\normtf{\bW}^2}.
  \end{align*}
\end{proof}

\begin{lemma}[Coupling between $f$ and $f^Q$]
  \label{lemma:coupling-f}
  We have for all $\bx\in\S^{d-1}(\Bx)$ that
  \begin{enumerate}[label=(\alph*)]
  \item $\E_\bSigma[f^L_{\bW\bSigma}(\bx)]=0$ and
    $\E_\bSigma[(f^L_{\bW\bSigma}(\bx))^2]\le
    \wt{O}(d^2\Bx^2\normtf{\bW}^2m^{-1/2})$.
  \item $|\Delta^Q_{\bW\bSigma}(\bx)|\le
    O(\Bx^3\normtf{\bW}^3m^{-1/4})$ (almost surely for all $\bSigma$.)
  \end{enumerate}
\end{lemma}
\begin{proof}
  \begin{enumerate}[label=(\alph*)]
    \item 
      Recall that
      \begin{equation*}
        f^L_{\bW\bSigma}(\bx) = \frac{1}{\sqrt{m}}\sum_{r\le m}
        a_r\sigma'(\bw_{0,r}^\top\bx)(\Sigma_{rr}\bw_r^\top\bx).
      \end{equation*}
      As $\Sigma_{rr}$ has mean zero, we have $\E_\bSigma[f^L]=0$ and
      \begin{align*}
        & \quad E_{\bSigma}[(f^L)^2] = \frac{1}{m}\sum_{r\le
          m}a_r^2\sigma'(\bw_{0,r}^\top\bx)^2(\bw_r^\top\bx)^2
        \\
        & \stackrel{(i)}{\le} \frac{1}{m}\sum_r
          C(\bw_{0,r}^\top\bx)^4(\bw_r^\top\bx)^2 \stackrel{(ii)}{\le}
          C\max_{r\in[m]}(\bw_{0,r}^\top\bx)^4 \cdot \frac{1}{m}\sum_r
          \Bx^2\ltwo{\bw_r}^2 \\ 
        & \stackrel{(iii)}{\le} \wt{O}(d^2\Bx^2) \cdot
          \frac{1}{m}\sum_{r\le m}\ltwo{\bw_r}^2
          \stackrel{(iv)}{\le} C\Bx^6\Bwzero^4 \cdot \left(
          \frac{1}{m}\sum_{r\le 
          m}\ltwo{\bw_r}^4 \right)^{1/2} =
          \wt{O}\left(d^2\Bx^2\normtf{\bW}^2m^{-1/2}\right).
      \end{align*}
      Above, (i) follows from the assumption that $|\sigma'(t)|\le
      Ct^2$, (ii) is Cauchy-Schwarz, (iii) uses the
      bound~\eqref{equation:w0-bound}, and (iv) uses the power mean
      inequality on $\ltwo{\bw_r}$.
    \item We have by the Lipschitzness of $\sigma''$ that
      \begin{align*}
        & \quad |\Delta^Q_{\bW}(\bx)| = \Big| \frac{1}{\sqrt{m}}\sum_{r\le
          m}a_r\Big(\sigma((\bw_{0,r}+\Sigma_{rr}\bw_r)^\top\bx) -
        \sigma(\bw_{0,r}^\top\bx) \\
        & \qquad\qquad\qquad -
          \sigma'(\bw_{0,r}^\top\bx)(\Sigma_{rr}\bw_r^\top\bx) -
          \sigma''(\bw_{0,r}^\top\bx)(\Sigma_{rr}\bw_r^\top\bx)^2
          \Big)\Big| \\
        & \le \frac{1}{\sqrt{m}} \sum_{r\le m}
          C|\Sigma_{rr}\bw_r^\top\bx|^3 \stackrel{(i)}{\le}
          C\sqrt{m}\Bx^3 \cdot \frac{1}{m}\sum_{r\le m}\ltwo{\bw_r}^3
        \\
        & \le C\sqrt{m}\Bx^3 \cdot \left(\frac{1}{m}\sum_{r\le
          m}\ltwo{\bw_r}^4\right)^{3/4} \\
        & = O\left( \Bx^3\normtf{\bW}^3m^{-1/4} \right),
      \end{align*}
      where again (i) uses the power mean inequality on $\ltwo{\bw_r}$.
  \end{enumerate}
\end{proof}

\subsection{Closeness of landscapes}
\begin{lemma}[$L^Q$ close to $L$]
  \label{lemma:coupling-l}
  We have for all $\bW\in\R^{d\times m}$ that
  \begin{align*}
    |L(\bW) - L^Q(\bW)| \le \wt{O}\paren{\Bx^3\normtf{\bW}^3m^{-1/4} +
      d^2\Bx^2\normtf{\bW}^2m^{-1/2}}.
  \end{align*}
\end{lemma}
\begin{proof}
  Recall that
  \begin{equation*}
    L(\bW) = \E_{\bSigma, \cD}[\ell(y,
    f_{\bW_0+\bW\bSigma}(\bx))]~~~{\rm and}~~~L^Q(\bW) =
    \E_{\cD}[\ell(y, f^Q_\bW(\bx))].
  \end{equation*}
  By the 1-Lipschitzness of $z\mapsto\ell(y,z)$ we have
  \begin{align*}
    & \quad \left|L(\bW) - L^Q(\bW)\right| \le
      \E_{\bSigma, \cD}\brac{\abs{f_{\bW_0+\bW\bSigma}(\bx) -
      f^Q_\bW(\bx)}} \\
    & \le \paren{\E_{\bSigma, \cD}\brac{ (f^L_{\bW\bSigma}(\bx) +
      \Delta^Q_{\bW\bSigma}(\bx))^2 }}^{1/2} \\
    & \le \paren{2 \E_{\bSigma, \cD}\brac{ (f^L_{\bW\bSigma}(\bx))^2}
      + 2\E_{\bSigma, \cD}\brac{(\Delta^Q_{\bW\bSigma}(\bx))^2}}^{1/2}
    \\
    & = \wt{O}\paren{ \Bx^3\normtf{\bW}^3m^{-1/4} +
      d^2\Bx^2\normtf{\bW}^2m^{-1/2}},
  \end{align*}
  where the last step uses Lemma~\ref{lemma:coupling-f}.
\end{proof}

\begin{lemma}[Closeness of directional gradients]
  \label{lemma:coupling-grad-l}
  We have
  \begin{align*}
    & \quad \abs{\<\grad L(\bW), \bW\> - \<\grad L^Q(\bW), \bW\>} \\
    & \le \wt{O}\paren{
      \paren{d\Bx\normtf{\bW} +
      \sqrt{d}\Bx^5\normtf{\bW}^5+\Bx^3\normtf{\bW}^3}m^{-1/4}
      + d^{2.5}\Bx^4\normtf{\bW}^4m^{-1/2}}.
  \end{align*}
\end{lemma}
\begin{proof}
  Differentiating $L$ and $L^Q$ and taking the inner
  product with $\bW$, we get
  \begin{equation*}
    \<\grad L(\bW), \bW\> = \E_{\bSigma,
      \cD} \brac{\ell'(y, f_{\bW_0+\bW\bSigma}(\bx)) \cdot
    \frac{1}{\sqrt{m}}\sum_{r\le m}a_r
    \sigma'((\bw_{0,r}+\Sigma_{rr}\bw_r)^\top\bx)(\Sigma_{rr}\bw_r^\top\bx)}, 
  \end{equation*}
  and
  \begin{equation*}
    \<\grad L^Q(\bW), \bW\> = \E_{\cD}\brac{\ell'(y, f^Q_\bW(\bx))\cdot
      \frac{1}{\sqrt{m}}\sum_{r\le m}a_r \sigma''(\bw_{0,r}^\top\bx)
      \cdot (\bw_r^\top\bx)^2}.
  \end{equation*}
  Therefore, by expanding
  $\sigma'((\bw_{0,r}+\Sigma_{rr}\bw_r)^\top\bx)$ and noticing that
  $\Sigma_{rr}^2\equiv 1$, we have
  \begin{align*}
    & \quad \abs{\<\grad L(\bW) - \grad L^Q(\bW), \bW\>}
      = \Bigg| \underbrace{\E_{\bSigma,
      \cD}\brac{\ell'(y, f_{\bW_0+\bW\bSigma}(\bx)) \cdot
      \frac{1}{\sqrt{m}}\sum_{r\le m}a_r 
      \sigma'(\bw_{0,r}^\top\bx)(\Sigma_{rr}\bw_r^\top\bx)}}_{\rm I} \\
    & \qquad + \underbrace{\E_{\bSigma,\cD}\brac{\paren{ \ell'(y,
      f_{\bW_0+\bW\bSigma}(\bx))  - \ell'(y, 
      f^Q_\bW(\bx))} \cdot \frac{1}{\sqrt{m}} \sum_{r\le m}a_r
      \sigma''(\bw_{0,r}^\top\bx) \cdot (\bw_r^\top\bx)^2
      }}_{\rm II} \\
    & \qquad +
      \E_{\bSigma,
      \cD}\Bigg[\ell'(y, f_{\bW_0+\bW\bSigma}(\bx)) \\
    & \qquad\qquad \underbrace{\cdot
      \frac{1}{\sqrt{m}}\sum_{r\le m}a_r 
      \paren{\sigma'((\bw_{0,r}+\Sigma_{rr}\bw_r)^\top\bx) -
      \sigma'(\bw_{0,r}^\top\bx) -
      \sigma''(\bw_{0,r}^\top\bx)(\Sigma_{rr}\bw_r^\top\bx)}
      (\Sigma_{rr}\bw_r^\top\bx)
      \Bigg]}_{\rm III} \Bigg|.
  \end{align*}
  We now bound the three terms separately. Recall that $|\ell'|\le 1$
  and $\ell'(y,z)$ is 1-Lipschitz in $z$. For term I we have by
  Cauchy-Schwarz that
  \begin{align*}
    & \quad \abs{\rm I} \le \paren{\E_{
      \cD}\brac{\frac{1}{m}\sum_{r\le m}
      a_r^2\sigma'(\bw_{0,r}^\top\bx)^2(\bw_r^\top\bx)^2}}^{1/2} \\
    & \le \paren{C\max_{r\in[m], i\in[n]} (\bw_{0,r}^\top\bx_i)^4
      \cdot \frac{1}{m}\sum_{r\le m}
       \ltwo{\bw_r}^2\Bx^2}^{1/2} \\
    & \le \wt{O}(d\Bx) \cdot \paren{\frac{1}{m}\sum_{r\le m}
      \ltwo{\bw_r}^4}^{1/4} = 
      O\paren{d\Bx \normtf{\bW} m^{-1/4}}.
  \end{align*}
  For term II, we have
  \begin{align*}
    & \quad \abs{\rm II}
      \stackrel{(i)}{\le} \paren{\E_{\bSigma,
      \cD}\brac{(f_{\bW_0+\bW\bSigma}(\bx) - f^Q_\bW(\bx))^2}}^{1/2}
      \cdot \paren{\E_{\cD}\brac{(f^Q_\bW(\bx))^2}}^{1/2} \\
    & \stackrel{(ii)}{\le} \wt{O}\paren{\Bx^3\normtf{\bW}^3m^{-1/4} +
      d^2\Bx^2\normtf{\bW}^2m^{-1/2}} \cdot
      \wt{O}(\sqrt{d}\Bx^2\normtf{\bW}^2) \\
    & = \wt{O}\paren{\sqrt{d}\Bx^5\normtf{\bW}^5m^{-1/4} +
      d^{2.5}\Bx^4\normtf{\bW}^4m^{-1/2}}.
  \end{align*}
  where (i) uses Cauchy-Schwarz and (ii) uses the bounds in
  Lemma~\ref{lemma:bound-fq} and~\ref{lemma:coupling-f}. For term III
  we first note by the smoothness of $\sigma'$ that
  \begin{align*}
    &\quad \abs{a_r \paren{\sigma'((\bw_{0,r}+\Sigma_{rr}\bw_r)^\top\bx) -
      \sigma'(\bw_{0,r}^\top\bx) -
      \sigma''(\bw_{0,r}^\top\bx)(\Sigma_{rr}\bw_r^\top\bx)}(\Sigma_{rr}\bw_r^\top\bx)} \\
    & \le C\abs{\Sigma_{rr}\bw_r^\top\bx}^3 \le C\Bx^3\ltwo{\bw_r}^3.
  \end{align*}
  Substituting this bound into term III yields
  \begin{align*}
    & \quad \abs{\rm III} \le 
      \frac{1}{\sqrt{m}}\sum_{r\le m} C\Bx^3\ltwo{\bw_r}^3
      \le C\sqrt{m}\Bx^3 \cdot
      \frac{1}{m}\sum_{r\le m}\ltwo{\bw_r}^3 \\
    & \le C\sqrt{m}\Bx^3 \cdot
      \paren{\frac{1}{m}\sum_{r\le m}\ltwo{\bw_r}^4}^{3/4} =
      O\paren{\Bx^3\normtf{\bW}^3m^{-1/4}}.
  \end{align*}
  Putting together the bounds for term I, II, III gives the desired result.
\end{proof}

\begin{lemma}[Closeness of Hessians]
  \label{lemma:coupling-gradsq-l}
  Let $\bSigma'$ denote a diagonal matrix with diagonal entries drawn
  i.i.d. from $\Unif\set{\pm 1}$.
  We have for all $\bW,\bW_\star\in\R^{d\times m}$ that
  \begin{align*}
    & \quad \abs{\E_{\bSigma'}\brac{\paren{\grad^2L(\bW) -
      \grad^2L^Q(\bW)}[\bW_\star\bSigma', \bW_\star\bSigma']}} \\
    & \le \wt{O}\Bigg(
      \paren{\Bx^3\normtf{\bW}+\sqrt{d}\Bx^5\normtf{\bW}^3}
      \normtf{\bW_\star}^2m^{-1/4} \\
    & \qquad + \paren{d^{2.5}\Bx^4\normtf{\bW}^2\normtf{\bW_\star}^2 
      + \Bx^2(d^2+\normti{\bW}^4\Bx^4)\normtf{\bW_\star}^2}m^{-1/2} \\
    & \qquad + d\Bx^4\normtf{\bW}^2\normtf{\bW_\star}^2m^{-1}
      \Bigg).
  \end{align*}
\end{lemma}
\begin{proof}
  Differentiating $L$ and $L^Q$ twice on the direction
  $\bW_\star\bSigma'$, we get
  \begin{align*}
    & \quad \grad^2L(\bW)[\bW_\star\bSigma', \bW_\star\bSigma'] \\
    & = \underbrace{\E_{\bSigma, \bSigma', \cD}\brac{\ell''(y,
      f_{\bW_0+\bW\bSigma}(\bx)) \cdot \paren{\frac{1}{\sqrt{m}}
      \sum_{r\le m} a_r\sigma'( (\bw_{0,r} + \Sigma_{rr}\bw_r)^\top \bx)
      (\Sigma_{rr}\Sigma'_{rr}\bw_{\star,r}^\top\bx)}^2}}_{{\rm I}(L)} \\
    & \qquad + \underbrace{\E_{\bSigma, \cD}\brac{\ell'(y, f_{\bW_0+\bW\bSigma}(\bx))
      \cdot \frac{1}{\sqrt{m}}\sum_{r\le m} a_r\sigma''(
      (\bw_{0,r}+\Sigma_{rr}\bw_r)^\top\bx)
      (\Sigma_{rr}\Sigma'_{rr}\bw_{\star,r}^\top\bx)^2}}_{{\rm II}(L)},
  \end{align*}
  and
  \begin{align*}
    & \quad \grad^2 L^Q(\bW)[\bW_\star\bSigma', \bW_\star\bSigma'] =
      \underbrace{\E_{\bSigma',\cD}\brac{\ell''(y, f^Q_\bW(\bx)) \cdot
      \paren{\frac{1}{\sqrt{m}} \sum_{r\le m} 
      a_r\sigma''(\bw_{0,r}^\top\bx)(\bw_r^\top\bx)(\Sigma'_{rr}\bw_{\star,r}^\top\bx)}^2}}_{{\rm I}(L^Q)}
    \\
    & \qquad + \underbrace{
      \E_{\bSigma',\cD}\brac{\ell'(y, f^Q_\bW(\bx)) \cdot
      \frac{1}{\sqrt{m}}\sum_{r\le
      m}a_r\sigma''(\bw_{0,r}^\top\bx)(\Sigma'_{rr}\bw_{\star,r}^\top\bx)^2}}_{{\rm
      II}(L^Q)} .
  \end{align*}
  We first bound the terms ${\rm I}(L)$ and ${\rm I}(L^Q)$. We have
  \begin{align*}
    & \quad {\rm I}(L) = 2\E_{\bSigma, \bSigma',\cD}\brac{
      \frac{1}{m}\sum_{r\le m} a_r^2\sigma'(
      (\bw_{0,r}+\Sigma_{rr}\bw_r)^\top\bx )^2
      (\bw_{\star,r}^\top\bx)^2} \\
    & \le C\cdot \sup_{\ltwo{\bx}=\Bx} \frac{1}{m}\sum_{r\le m}
      \paren{(\bw_{0,r}+\Sigma_{rr}\bw_r)^\top\bx}^4
      (\bw_{\star,r}^\top\bx)^2 \\
    & \le C\Bx^2\cdot \frac{1}{m}\sum_{r\le m}(\wt{O}(d^2)+
      \ltwo{\bw_r}^4\Bx^4)\ltwo{\bw_{\star,r}}^2 \\
    & \le \wt{O}\paren{\Bx^2\paren{d^2 +
      \normti{\bW}^4\Bx^4}\normtf{\bW_\star}^2m^{-1/2}}.
  \end{align*}
  Using similar arguments on ${\rm I}(L^Q)$ gives the bound
  \begin{align}
    \label{equation:lq-vanishing}
    {\rm I}(L^Q) \le
    \wt{O}\paren{d\Bx^4\normtf{\bW}^2\normtf{\bW_\star}^2m^{-1}}.
  \end{align}
  We now shift attention to bounding ${\rm II}(L)-{\rm
    II}(L^Q)$. First note that
  \begin{align*}
    & \quad \abs{\Delta^{\sigma''}_r(\bx)} \defeq
      \abs{a_r(\sigma''((\bw_{0,r}+\Sigma_{rr}\bw_r)^\top\bx) -
      \sigma''(\bw_{0,r}^\top\bx))(\bw_{\star,r}^\top\bx)^2} \\
    & \le C|\Sigma_{rr}\bw_r^\top\bx| \cdot (\bw_{\star,r}^\top\bx)^2
      \le C\Bx^3\ltwo{\bw_r}\ltwo{\bw_{\star,r}}^2.
  \end{align*}
  Then we have, by applying the bounds in Lemma~\ref{lemma:bound-fq}
  and~\ref{lemma:coupling-f}, 
  \begin{align*}
    & \quad \abs{{\rm II}(L) - {\rm II}(L^Q)} \\
    & = \abs{\E_{\bSigma,
      \cD}\brac{\ell'(y, f_{\bW_0+\bW\bSigma}(\bx)) \cdot
      \frac{1}{\sqrt{m}}\sum_{r\le m}\Delta^{\sigma''}_r(\bx)} +
      \E_{\bSigma, \cD}\brac{\paren{\ell'(y, f^Q_{\bW_0+\bW\bSigma}(\bx)) -
      \ell'(y, f^Q_\bW(\bx))}
      \cdot 2f^Q_{\bW_\star}(\bx)}} \\
    & \le C \cdot \frac{1}{\sqrt{m}}\sum_{r\le
      m}\Bx^3\ltwo{\bw_r}\ltwo{\bw_{\star,r}}^2 +
      C\sqrt{\E\brac{\paren{ f^L_{\bW\bSigma}(\bx) +
      \Delta^Q_{\bW\bSigma}(\bx)  }^2}} \cdot
      \paren{\E\brac{f^Q_{\bW_\star}(\bx)^2}}^{1/2} \\
    & \le 
      \wt{O}\paren{\Bx^3\normtf{\bW}\normtf{\bW_\star}^2 m^{-1/4}}
      +  \wt{O}\paren{\Bx^3\normtf{\bW}^3m^{-1/4} +
      d^2\Bx^2\normtf{\bW}^2m^{-1/2}} \cdot
      \wt{O}\paren{\sqrt{d}\Bx^2\normtf{\bW_\star}^2} .\\
    & =
      \wt{O}\paren{
      \paren{\Bx^3\normtf{\bW}+\sqrt{d}\Bx^5\normtf{\bW}^3}
      \normtf{\bW_\star}^2m^{-1/4}
      + d^{2.5}\Bx^4\normtf{\bW}^2\normtf{\bW_\star}^2m^{-1/2}
      }.
  \end{align*}
  Combining all the bounds gives the desired result.
\end{proof}

\subsection{Proof of Theorem~\ref{theorem:good-landscape-l}}
\label{appendix:proof-good-landscape-l}
We apply Lemma~\ref{lemma:coupling-l},~\ref{lemma:coupling-grad-l},
and~\ref{lemma:coupling-gradsq-l} to connect the neural net loss $L$
to the ``clean risk'' $L^Q$. First, by
Lemma~\ref{lemma:coupling-l}, we have for all the assumed $\bW$ that
\begin{equation*}
  \abs{L(\bW) - L^Q(\bW)} \le
  \wt{O}\paren{\Bx^3\normtf{\bW}^3m^{-1/4} + 
    d^2\Bx^2\normtf{\bW}^2m^{-1/2}}.
\end{equation*}
Therefore we have $\abs{L(\bW) - L^Q(\bW)}\le \eps/6$
so long as
\begin{equation}
  \label{equation:large-m-from-l}
  m \ge \wt{O}\paren{\Bx^{12}\Bw^{12}\eps^{-4} +
    d^4\Bx^4\Bw^4\eps^{-2}}.
\end{equation}
Applying Lemma~\ref{lemma:clean-risk}, we obtain that
\begin{equation}
  \label{equation:lq-landscape}
  \begin{aligned}
    & \quad \E_{\bSigma'}\brac{\grad^2 L^Q(\bW)[\bW_\star\bSigma', 
      \bW_\star\bSigma']} - \<\grad L^Q(\bW), \bW\> \\
    & \le 2(L^Q(\bW) - \opt) + \eps/3 \le 2(L(\bW) - \opt) + 2\eps/3
  \end{aligned}
\end{equation}
provided that the error term in Lemma~\ref{lemma:clean-risk} is
bounded by $\eps/3$, which happens when
\begin{equation}
  \label{equation:large-m-from-lq}
  m \ge \wt{O}\paren{d\Bx^4\Bw^2\Bwstar^2\eps^{-1}}.
\end{equation}

Finally, we choose $m$ sufficiently large so that
\begin{align*}
  \abs{\E_{\bSigma'}\brac{(\grad^2 L(\bW) -
  \grad^2L^Q(\bW))[\bW_\star\bSigma', 
  \bW_\star\bSigma']}} \le \eps/6
\end{align*}
and
\begin{align*}
  \abs{\<\grad L(\bW) - \grad L^Q(\bW), \bW\>} \le \eps/6,
\end{align*}
which combined with~\eqref{equation:lq-landscape} yields the
desired result. By Lemma~\ref{lemma:coupling-grad-l}
and~\ref{lemma:coupling-gradsq-l}, it suffices to choose $m$ such
that, to satisfy the closeness of directional gradients,
\begin{align}
  \label{equation:large-m-from-grad-l}
  m \ge \wt{O}\paren{
  \paren{d^4\Bx^4\Bw^4 + d^2\Bx^{20}\Bw^{20} +
  \Bx^{12}\Bw^{12}}\eps^{-4}
  + d^5\Bx^8\Bw^8\eps^{-2}},
\end{align}
and to satisfy the closeness of Hessian quadratic forms,
\begin{equation}
  \label{equation:large-m-from-gradsq-l}
  \begin{aligned}
    & m \ge O\bigg(
    \brac{\Bx^{12}\Bw^4\Bwstar^8 + d^2\Bx^{20}\Bw^{12}\Bwstar^8}
    \eps^{-4} \\ 
    & \qquad\qquad
    + \brac{d^5\Bx^8\Bw^4\Bwstar^4 + d^4\Bx^4\Bwstar^4 +
      \Bx^{12}\Bw^8\Bwstar^4} \eps^{-2}
    + d\Bx^4\Bw^2\Bwstar^2\eps^{-1}\bigg).
  \end{aligned}
\end{equation}
Collecting the requirements on $m$
in~\eqref{equation:large-m-from-l},~\eqref{equation:large-m-from-lq},~\eqref{equation:large-m-from-grad-l},~\eqref{equation:large-m-from-gradsq-l}
and merging terms using $\eps\le 1$ and $\Bwstar\le \Bw$, the desired result holds whenever
\begin{align*}
  m \ge O\paren{
  \brac{\Bx^{12}\Bw^{12} + d^4\Bx^4\Bw^4 +
  d^2\Bx^{20}\Bw^{20}}\eps^{-4}
  + d^5\Bx^8\Bw^8\eps^{-2}}.
\end{align*}
This completes the proof. \qed

\subsection{Proof of Corollary~\ref{corollary:good-landscape-l-lambda}}
\label{appendix:proof-good-landscape-l-lambda}
\begin{proof}
  For all $\lambda\ge 0$ define
  \begin{equation*}
    A_\lambda \defeq \E_{\bSigma'}\brac{\grad^2
      L_\lambda(\bW)[\bW_\star\bSigma', 
      \bW_\star\bSigma']} - \<\grad L_\lambda(\bW), \bW\> +
    2(L_\lambda(\bW) - \opt),
  \end{equation*}
  By Lemma~\ref{theorem:good-landscape-l}, it suffices to show that
  \begin{equation*}
    A_\lambda - A_0 \le C\lambda\normtf{\bW_\star}^8 -
    \lambda\normtf{\bW}^8 
  \end{equation*}
  for some absolute constant $C$.
  
  Recall that $L_\lambda(\bW)=L(\bW) +
  \lambda\normtf{\bW_\star}^8$. By differentiating
  $\bA\mapsto\normtf{\bA}^8$ we get
  \begin{align*}
    \<\grad (L_\lambda - L)(\bW), \bW\> = 2\normtf{\bW}^4 \cdot
    \sum_{r\le m} \<4\lambda\ltwo{\bw_r}^2\bw_r, \bw_r\> =
    8\lambda\normtf{\bW}^8
  \end{align*}
  and
  \begin{align*}
    & \quad
      \grad^2 (L_\lambda - L)(\bW)[\bW_\star\bSigma',
      \bW_\star\bSigma'] \\
    & = 8\lambda\normtf{\bW}^4\brac{\sum_{r\le m}
      \ltwo{\bw_r}^2\ltwo{\bw_{\star,r}\Sigma'_{rr}}^2 +
      2\<\bw_r, \bw_{\star,r}\Sigma'_{rr}\>^2} +
      32\lambda\normtf{\bW}^4 \cdot \paren{\sum_{r\le m}\ltwo{\bw_r}^2
      \<\bw_r, \bw_{\star,r}\Sigma'_{rr}\>}^2
    \\
    & \le 56\lambda\normtf{\bW}^4 \sum_{r\le m}
      \ltwo{\bw_r}^2\ltwo{\bw_{\star,r}}^2
      \stackrel{(i)}{\le} 56\lambda\normtf{\bW}^6\normtf{\bW_\star}^2
      \stackrel{(ii)}{\le} 14\lambda\alpha\normtf{\bW}^8 +
      \frac{378\lambda}{\alpha^3}\normtf{\bW_\star}^8,
  \end{align*}
  where (i) used Cauchy-Schwarz and (ii) used the AM-GM inequality
  $p^3q\le \alpha p^4/4+27q^4/(4\alpha^3)$ for all $p,q$ and
  $\alpha> 0$. Substituting the above expressions into
  $A_\lambda - A_0$ yields
  \begin{align*}
    & \quad A_\lambda - A_0 \\
    & \le 14\lambda\alpha\normtf{\bW}^8 +
    \frac{378\lambda}{\alpha^3}\normtf{\bW_\star}^8 -
      8\lambda\normtf{\bW}^8 
      + 2\lambda\normtf{\bW}^8 \\
    & = (14\lambda\alpha - 6\lambda)\normtf{\bW}^8 +
    \frac{378\lambda}{\alpha^3}\normtf{\bW_\star}^8.
  \end{align*}
  Choosing $\alpha=5/14$ gives the desired result.
\end{proof}


\subsection{Proof of Theorem~\ref{theorem:main}}
\label{appendix:proof-main}
We begin by choosing the regularization strength as
\begin{equation*}
  \lambda = \lambda_0\Bwstar^{-8},
\end{equation*}
where $\lambda_0$ is a constant to be determined. Let $\eps$ be an
accuracy parameter also to be determined.

\paragraph{Localizing second-order stationary points}
We first argue that any second order stationary point $\bW$ has to
satisfy $\normtf{\bW}\le \Bwlarge$ for some large but controlled
$\Bwlarge$. We first note that for the clean risk $L^Q$, we have for
any $\bW\in\R^{d\times m}$ that
\begin{align*}
  & \quad \<\grad L^Q(\bW), \bW\> = \E_{\cD}\brac{\ell'(y,
    f^Q_{\bW}(\bx)) \cdot 2f^Q_{\bW}(\bx)} \\
  & = 2\E_{\cD}\brac{\ell'(y, f^Q_{\bW}(\bx))
    \cdot (f^Q_{\bW}(\bx) - f^Q_{\bzero}(\bx))} \stackrel{(i)}{\ge}
    2(L^Q(\bW) - L^Q(\bzero)) \stackrel{(ii)}{\ge} -2,
\end{align*}
where (i) uses convexity of $\ell$ and (ii) uses the assumption that
$\ell(y, 0)\le 1$ for all $y\in\mc{Y}$.

Now, applying the coupling
Lemma~\ref{lemma:coupling-grad-l}, and
combining with the fact that $\<\grad_\bW (\lambda\normtf{\bW}^8),
\bW\>=8\lambda\normtf{\bW}^8$, we have simultaneously for all $\bW$
that
\begin{align*}
  & \quad \<\grad L_\lambda(\bW), \bW\> \\
  & \ge \<\grad_\bW (\lambda\normtf{\bW}^8),
    \bW\> + \<\grad L^Q(\bW), \bW\> - \abs{\<\grad (L - L^Q)(\bW),
    \bW\>} \\
  & \ge 8\lambda\normtf{\bW}^8 -2
    - \wt{O}\paren{
  \paren{d\Bx\normtf{\bW} +
  \sqrt{d}\Bx^5\normtf{\bW}^5+\Bx^3\normtf{\bW}^3}m^{-1/4}
  + d^{2.5}\Bx^4\normtf{\bW}^4m^{-1/2}}.
\end{align*}
Therefore we see that any stationary point $\bW$ has to satisfy
\begin{align*}
  & \quad \normtf{\bW} \le \Bwlarge \\
  & \defeq \wt{O}\paren{\lambda^{-1/8} +
    (\lambda^{-1}d\Bx m^{-1/4})^{1/7} +
    (\lambda^{-1}\sqrt{d}\Bx^5m^{-1/4})^{1/3}  +
    (\lambda^{-1}\Bx^3)^{1/5} +
    (\lambda^{-1}d^{2.5}\Bx^4m^{-1/2})^{1/4}}.
\end{align*}
By Corollary~\ref{corollary:good-landscape-l-lambda}, choosing $m\ge
{\rm poly}(\lambda_0^{-1}, d, \Bwstar\Bx, \eps)$, the coupling error
is bounded by $\eps$ in $\balltf(\Bwlarge)$, i.e. for all
$\bW\in\balltf(\Bwlarge)$ we have that
\begin{equation}
  \label{equation:negative-lambda-bound}
  \begin{aligned}
    & \quad \E_{\bSigma'}\brac{\grad^2
      L_\lambda(\bW)[\bW_\star\bSigma', 
      \bW_\star\bSigma']} \\
    & \le \<\grad L_\lambda(\bW), \bW\> -
      2(L_\lambda(\bW) - \opt) - \lambda\normtf{\bW}^8 +
      C\lambda\normtf{\bW_\star}^8 + \eps,
  \end{aligned}
\end{equation}
where $C=O(1)$ is an absolute constant.

\paragraph{Bounding loss and norm}
Choosing
\begin{equation*}
  \lambda_0 = \frac{1}{C} \cdot (2\gamma\opt + \eps),
\end{equation*}
we get that
$C\lambda\Bwstar^8=2\gamma\opt + \eps$,
and thus the bound~\eqref{equation:negative-lambda-bound} reads 
\begin{align*}
    & \quad \E_{\bSigma'}\brac{\grad^2
      L_\lambda(\bW)[\bW_\star\bSigma', 
      \bW_\star\bSigma']} \\
    & \le \<\grad L_\lambda(\bW), \bW\> -
      2(L_\lambda(\bW) - \opt) - \lambda\normtf{\bW}^8 + 2\gamma\opt + 
      2\eps. 
\end{align*}
For the second-order stationary point $\what{\bW}$, the gradient term
vanishes and the Hessian term is non-negative, so we get
\begin{equation*}
  2(L_\lambda(\what{\bW}) - \opt) \le 2(\gamma\opt + \eps) -
  \lambda\normtf{\what{\bW}}^8 \le 2(\gamma\opt + \eps)
\end{equation*}
and thus
\begin{equation*}
  L_\lambda(\what{\bW}) \le (1+\gamma)\opt + \eps.
\end{equation*}
Further, by re-writing~\eqref{equation:negative-lambda-bound}, we
obtain
\begin{align*}
  & \quad \lambda\normtf{\what{\bW}}^8 \le C\lambda\Bwstar^8 + 2(\opt -
    L_\lambda(\what{\bW})) + \eps \le C\lambda\Bwstar^8 + 2\opt + \eps
  \\
  & \le C\lambda\Bwstar^8 \cdot \paren{1 + \frac{2\opt +
    \eps}{2\gamma\opt + \eps}} = O(1) \cdot \lambda\Bwstar^8,
\end{align*}
for any $\gamma=O(1)$. This is the desired result. \qed

\section{Proofs for Section~\ref{section:gen}}
\subsection{Proof of Lemma~\ref{lemma:gen-opnorm}}
\label{appendix:proof-gen-opnorm}
As the loss $\ell(y,z)$ is 1-Lipschitz in $z$ for all $y$, by the
Rademacher contraction theorem~\citep[Chapter 5]{wainwright2019high}
we have that
\begin{align*}
  & \quad \E_{\bsigma, \bx}\brac{\sup_{\normtf{\bW}\le \Bw}
    \frac{1}{n}\sum_{i=1}^n \sigma_i\ell(y_i, f^Q_\bW(\bx_i))} \\
  & \le 2\E_{\bsigma,
    \bx}\brac{\sup_{\normtf{\bW}\le \Bw} \frac{1}{n}\sum_{i=1}^n
    \sigma_i f^Q_\bW(\bx_i)} + \E_{\bsigma,
    \bx}\brac{\frac{1}{n}\sum_{i=1}^n
    \sigma_i\ell(y_i, 0)} \\
  & \le \E_{\bsigma,\bx}\brac{\sup_{\normtf{\bW}\le \Bw}
    \frac{1}{\sqrt{m}}\sum_{r\le m} 
    \<\frac{1}{n}\sum_{i=1}^n\sigma_i
    a_r\sigma''(\bw_{0,r}^\top\bx_i)\bx_i\bx_i^\top,
    \bw_r\bw_r^\top\>} + \frac{1}{\sqrt{n}} \\
  & \le \E_{\bsigma,
    \bx}\brac{\sup_{\normtf{\bW}\le \Bw}
    \max_{r\in[m]}\opnorm{\frac{1}{n}\sum_{i=1}^n
    a_r\sigma_i\sigma''(\bw_{0,r}^\top\bx_i)\bx_i\bx_i^\top} \cdot
    \frac{1}{\sqrt{m}}\sum_{r\le m}\lnuc{\bw_r\bw_r^\top}} +
    \frac{1}{\sqrt{n}}\\ 
  & \le \E_{\bsigma,\bx}\brac{\max_{r\in[m]}\opnorm{\frac{1}{n}\sum_{i=1}^n
    \sigma_i\sigma''(\bw_{0,r}^\top\bx_i)\bx_i\bx_i^\top}} \cdot
    \underbrace{\sup_{\normtf{\bW}\le \Bw}
    \frac{1}{\sqrt{m}}\sum_{r\le m}\ltwo{\bw_r}^2}_{\le \Bw^2} +
    \frac{1}{\sqrt{n}},
\end{align*}
where the last step used the power mean (or Cauchy-Schwarz)
inequality on $\set{\ltwo{\bw_r}}$. \qed

\subsection{Proof of Theorem~\ref{theorem:gen-main}}
\label{appendix:proof-gen-main}
We first relate the generalization of $L$ to that of $L^Q$ through
\begin{equation*}
  L_P(\what{\bW}) - L(\what{\bW}) \le L_P(\what{\bW}) - L^Q_P(\what{\bW})
  + L^Q_P(\what{\bW}) - L^Q(\what{\bW}) + L^Q(\what{\bW}) -
  L(\what{\bW}).
\end{equation*}
By Lemma~\ref{lemma:coupling-l}, we have simultaneously for all
$\bW\in\balltf(\Bw)$ that
\begin{equation}
  \label{equation:gen-coupling-err-1}
  \abs{L(\bW) - L^Q(\bW)} \le \wt{O}\paren{\Bx^3\Bw^3m^{-1/4}
    + d^2\Bx^2\Bw^2m^{-1/2}}.
\end{equation}
Further, from the proof we see that the argument does not depend on
the distribution of $\bx$ (it holds uniformly for all
$\bx\in\S^{d-1}(\Bx)$, therefore for the population version we also
have the bound
\begin{equation}
  \label{equation:gen-coupling-err-2}
  \abs{L_P(\bW) - L^Q_P(\bW)} \le
  \wt{O}\paren{\Bx^3\Bw^3m^{-1/4} 
    + d^2\Bx^2\Bw^2m^{-1/2}}.
\end{equation}
These bounds hold for all $\bW\in\balltf(\Bw)$ so apply to
$\what{\bW}$. Therefore it remains to bound $L^Q_P(\what{\bW}) -
L^Q(\what{\bW})$, i.e. the generalization of the quadratic model.

\paragraph{Generalization of quadratic model}
By symmetrization and applying Lemma~\ref{lemma:gen-opnorm}, we have
\begin{equation}
  \label{equation:gen-to-opnorm}
\begin{aligned}
  & \quad \E_{\bW_0,\cD}\brac{L^Q_P(\what{\bW}) - L^Q(\what{\bW})}
  \le \E_{\bW_0,\cD}\brac{\sup_{\normtf{\bW}\le \Bw} L^Q_P(\bW) - L^Q(\bW)}
  \\
  & \le 2\E_{\bW_0,\bsigma, \bx}\brac{\sup_{\normtf{\bW}\le \Bw}
    \frac{1}{n}\sum_{i=1}^n \sigma_i\ell(y_i, f^Q_\bW(\bx_i))} \\
  & \le 4\Bw^2\E_{\bW_0,\bsigma, \bx} \brac{ \max_{r\in[m]}
    \opnorm{\frac{1}{n}\sum_{i=1}^n 
    \sigma_i\sigma''(\bw_{0,r}^\top\bx_i)\bx_i\bx_i^\top}} +
    \frac{2}{\sqrt{n}}.
  \end{aligned}
\end{equation}
We now focus on bounding the expected max operator norm above. First,
we apply the matrix concentration
Lemma~\ref{lemma:matrix-concentration} to deduce that
\begin{align*}
  & \quad \E_{\bW_0,\bsigma,\bx}\brac{ \max_{r\in[m]}
    \opnorm{\frac{1}{n}\sum_{i=1}^n 
    \sigma_i\sigma''(\bw_{0,r}^\top\bx_i)\bx_i\bx_i^\top} } \\
  & \le 4\sqrt{\log(2dm)} \cdot
  \E_{\bW_0, \bx}\brac{\sqrt{\max_{r\in[m]}
  \opnorm{\frac{1}{n^2}\sum_{i=1}^n
    \sigma''(\bw_{0,r}^\top\bx_i)^2\ltwo{\bx_i}^2\bx_i\bx_i^\top}}} \\
  & \le 4\Bx\sqrt{\frac{\log(2dm)}{n}} \cdot \E_{\bW_0,\bx}\brac{
    \sqrt{\max_{r,i}\sigma''(\bw_{0,r}^\top\bx_i)^2  \cdot
    \opnorm{\frac{1}{n}\sum_{i=1}^n \bx_i\bx_i^\top}}} \\
  & \le 4\Bx\sqrt{\frac{\log(2dm)}{n}}
    \paren{\E_{\bW_0,\bx}\brac{\max_{r,i}\sigma''(\bw_{0,r}^\top\bx_i)^2}
    \cdot
    \E_{\bx}\brac{\opnorm{\frac{1}{n}\sum_{i=1}^n\bx_i\bx_i^\top}}}^{1/2}
\end{align*}
As $|\sigma''(t)|\le Ct$ and 
$\bw_{0,r}^\top\bx_i\sim \normal(0, 1)$ for all $(r,i)$, by standard
expected max bound on sub-exponential variables we have
\begin{equation*}
  \E_{\bW_0,\bx}\brac{\max \sigma''(\bw_{0,r}^\top\bx_i)^2} \le
  O\paren{\log(mn)} = \wt{O}(1).
\end{equation*}
Therefore defining
\begin{equation*}
  \Mxop \defeq \paren{\Bx^{-2} \cdot
    \E_{\bx}\brac{\frac{1}{n}\sum_{i=1}^n \bx_i\bx_i^\top}}^{1/2},
\end{equation*}
and substituting the above bound into~\eqref{equation:gen-to-opnorm}
yields that
\begin{equation*}
  \E_{\bW_0,\cD}\brac{L^Q_P(\what{\bW}) - L^Q(\what{\bW})} \le
  \wt{O}\paren{\frac{\Bx^2\Bw^2\Mxop}{\sqrt{n}} +
    \frac{1}{\sqrt{n}}}.
\end{equation*}
Combining the bound with the
coupling error~\eqref{equation:gen-coupling-err-1}
and~\eqref{equation:gen-coupling-err-2}, we arrive at the desired
result.

For $\Mxop$ we have two versions of bounds:
\begin{enumerate}[label=(\alph*)]
\item We always have $\opnorm{\sum_{i\le n}\bx_i\bx_i^\top/n}\le
  \Bx^2$, and thus $\Mxop\le 1$.
\item If, in addition, $\bx$ is uniformly distributed on the sphere
  $\S^{d-1}(\Bx)$ or the hypercube $\set{\pm \Bx/\sqrt{d}}^d$, then we
  have by standard covariance concentration~\citep[Theorem
  4.7.1]{vershynin2018high} that
  $\E_{\bx}\brac{\opnorm{\sum_{i\le n}\bx_i\bx_i^\top / n}} \le
  \Bx^2/d \cdot O(1+\sqrt{d/n}+d/n)=O(\Bx^2/d)$ when $n\ge d$.
  More generally, if for all $\bv\in\S^{d-1}(1)$ we have
  \begin{equation*}
    \norm{\bv^\top\bx}_{\psi_2} \le K\sqrt{\bv^\top\Cov(\bx)\bv},
  \end{equation*}
  and that $\kappa(\Cov(\bx)) \le \kappa$, then we have
  $\opnorm{\Cov(\bx))}\le \kappa\Bx^2/d$. Applying~\citep[Theorem
  4.7.1]{vershynin2018high}, we get $\Mxop\le \kappa/\sqrt{d}$ whenever
  $n\ge O(K^4d)$.
\end{enumerate}
\qed

\subsection{Expressive power of infinitely wide quadratic models}
\begin{lemma}[Expressivity of $f^Q$ with infinitely many neurons]
  \label{lemma:quad-net-expressivity-infinite-m}
  Suppose
  $f_\star(\bx) = \alpha(\bbeta^\top \bx)^p$ for some $\alpha\in\R$,
  $\bbeta\in\R^d$, and $p\ge 2$ and such that
  $p-2\in\set{1}\cup\set{2\ell}_{\ell\ge 0}$. Suppose further that we
  use $\sigma(t) = \frac{1}{6}\relu^3(t)$ (so that $\sigma''(t) =
  \relu(t)$), then there exists choices of $(\bw_+, \bw_-)$ that
  depends on $\bw_0$ such that
  \begin{equation*}
    \E_{\bw_0}\brac{ \sigma''(\bw_0^\top \bx)
      \paren{(\bw_+^\top\bx)^2 - (\bw_-^\top\bx)^2}} = f_\star(\bx)
  \end{equation*}
  and further satisfies the norm bound
  \begin{equation*}
    \E_{\bw_0}\brac{\ltwo{\bw_+}^4 + \ltwo{\bw_-}^4} \le 2\pi ((p-2)\vee
    1)^3\alpha^2 \Bx^{2(p-2)}\ltwo{\bbeta}^{2p}.
  \end{equation*}
\end{lemma}
\begin{proof}
  Our proof builds on reducing the problem from representing
  $(\bbeta^\top\bx)^p$ via quadratic networks to representing
  $(\bbeta^\top\bx)^{p-2}$ through a random feature
  model. More precisely, we consider choosing
  \begin{equation}
    \label{equation:coupled-quad-w}
    (\bw_+, \bw_-) = \paren{\sqrt{[a]_+}\cdot \bbeta, \sqrt{[a]_-}\cdot
    \bbeta},
  \end{equation}
  where $a$ is a real-valued random scalar that can depend on $\bw_0$,
  and $\bbeta$ is the fixed coefficient vector in $f_\star$. With this
  choice, the quadratic network reduces to
  \begin{align*}
    & \quad \E_{\bw_0}\brac{ \sigma''(\bw_0^\top \bx)
      \paren{(\bw_+^\top\bx)^2 - (\bw_-^\top\bx)^2}} \\
    & = \E_{\bw_0}\brac{ \sigma''(\bw_0^\top \bx) 
      \paren{a_+(\bbeta^\top\bx)^2 - a_-(\bbeta^\top\bx)^2}} =
      (\beta^\top\bx)^2 \E_{\bw_0}\brac{ \sigma''(\bw_0^\top \bx) a}. 
  \end{align*}
  Therefore, to let the above express
  $f_\star(\bx)=\alpha(\bbeta^\top\bx)^p$, it suffices to choose $a$
  such that 
  \begin{equation}
    \label{equation:express-pminus2}
    \E[\sigma''(\bw_0^\top \bx) a] \equiv \alpha(\bbeta^\top \bx)^{p-2}
  \end{equation}
  for all $\bx$. By Lemma~\ref{lemma:relu-rf-expressivity}, there
  exists $a=a(\bw_0)$ satisfying~\eqref{equation:express-pminus2} and
  such that 
  \begin{equation*}
    \E_{\bw_0}[a^2] \le 2\pi ((p-2)\vee
    1)^3\alpha^2 \Bx^{2(p-2)}\ltwo{\bbeta}^{2(p-2)}.
  \end{equation*}
  Using this $a$ in~\eqref{equation:coupled-quad-w}, the quadratic
  network induced by $(\bw_+,\bw_-)$ has the desired expressivity, and
  further satisfies the expected 4th power norm bound
  \begin{align*}
    & \quad \E_{\bw_0}[\ltwo{\bw_+}^4 + \ltwo{\bw_-}^4] \\
    & = \E_{\bw_0}[[a]_+^2 +
      [a]_-^2] \cdot \ltwo{\bbeta}^4 = \E_{\bw_0}[a^2]\ltwo{\bbeta}^4
      \le 2\pi ((p-2)\vee
    1)^3\alpha^2 \Bx^{2(p-2)}\ltwo{\bbeta}^{2p}.
  \end{align*}
  This is the desired result.

\end{proof}

\subsection{Proof of
  Theorem~\ref{theorem:quad-net-expressivity-sum}}
\label{appendix:proof-quad-net-expressivity-sum}

We begin by stating and proving the result for $k=1$ in
Appendix~\ref{appendix:expressivity-single}, i.e. when
$f_\star=\alpha(\bbeta^\top\bx)^p$ is a single ``one-directional''
polynomial. The main theorem then follows as a straightforward extension of
the $k=1$ case, which we prove in
Appendix~\ref{appendix:expressivity-multiple}. 

\subsubsection{Expressing a single ``one-directional''
  polynomial}
\label{appendix:expressivity-single}
\begin{theorem}[Expressivity of $f^Q$]
  \label{theorem:quad-net-expressivity}
  Suppose $\set{(a_r, \bw_{0,r})}$ are generated according to the
  symmetric initialization~\eqref{equation:symmetric-init}, and
  $f_\star(\bx)=\alpha(\bbeta^\top\bx)^p$ where
  $p-2\in\set{1}\cup\set{2\ell}_{\ell\ge 0}$. Suppose further that we
  use $\sigma(t)=\frac{1}{6}\relu^3(t)$ (so that
  $\sigma''(t)=\relu(t)$), then so long as the width is sufficiently
  large:
  \begin{equation*}
    m\ge
    \wt{O}\paren{ndp^3\alpha^2(\Bx\ltwo{\bbeta})^{2p}\eps^{-2}},
  \end{equation*}
  we have with probability at least $1-\delta$ (over $\bW_0$) that
  there exists $\bW_\star\in\R^{d\times m}$ such that 
  \begin{equation*}
    \abs{L^Q(\bW_\star) - L(f_\star)} \le \eps~~~{\rm and}~~~
    \normtf{\bW_\star}^4 \le \Bwstar^4 =
    O\paren{p^3\alpha^2\Bx^{2(p-2)}\ltwo{\bbeta}^{2p}\delta^{-1}}. 
  \end{equation*}
\end{theorem}

\begin{proof-of-theorem}[\ref{theorem:quad-net-expressivity}]
We build on the infinite-neuron construction in
Lemma~\ref{lemma:quad-net-expressivity-infinite-m}. Given the
symmetric initialization $\set{\bw_{0,r}}_{r=1}^m$, for all
$r\in[m/2]$, we consider $\bW_\star\in\R^{d\times m}$ defined through
\begin{equation*}
  (\bw_{\star,r}, \bw_{\star,r+m/2}) =
  \paren{2m^{-1/4}\bw_+(\bw_{0,r}),
    2m^{-1/4}\bw_-(\bw_{0,r})},
\end{equation*}
where we recall $(\bw_+(\bw_0), \bw_-(\bw_0)) =
(\sqrt{a_+(\bw_0)}\bbeta, \sqrt{a_-(\bw_0)}\bbeta)$.
We then have
\begin{align*}
  & \quad f^Q_{\bW_\star}(\bx) = \frac{1}{2\sqrt{m}}\sum_{r\le
    m/2}\sigma''(\bw_{0,r}^\top\bx) \brac{(\bw_{\star,r}^\top\bx)^2 -
    (\bw_{\star,r+m/2}^\top\bx)^2} \\
  & = \frac{2}{m}\sum_{r\le m/2} \sigma''(\bw_{0,r}^\top\bx) \brac{
    (\bw_+(\bw_{0,r})^\top\bx)^2 - (\bw_-(\bw_{0,r})^\top\bx)^2 } \\
  & = \brac{\frac{1}{m/2}\sum_{r\le m/2} \sigma''(\bw_{0,r}^\top\bx)
    a(\bw_{0,r})} \cdot (\bbeta^\top\bx)^2.
\end{align*}

\paragraph{Bound on $\normtf{\bW_\star}$}
As $f_\star(\bx)=\alpha(\bbeta^\top\bx)^p$,
Lemma~\ref{lemma:quad-net-expressivity-infinite-m} guarantees that the
coefficient $a(\bw_0)$ involved above satisfies that
\begin{equation*}
  R_a^2 \defeq \E_{\bw_0}[a(\bw_0)^2] \le 2\pi((p-2)\vee
  1)^3\alpha^2\Bx^{2(p-2)}\ltwo{\bbeta}^{2(p-2)}.
\end{equation*}
By Markov inequality, we have with probability at least $1-\delta/2$
that
\begin{equation*}
  \frac{1}{m/2}\sum_{r\le m} a(\bw_{0,r})^2 \le 4\pi((p-2)\vee
  1)^3\alpha^2\Bx^{2(p-2)}\ltwo{\bbeta}^{2(p-2)}\delta^{-1},
\end{equation*}
which yields the bound
\begin{align*}
  & \quad \normtf{\bW}^4 = \sum_{r\le m}\ltwo{\bw_{\star, r}}^4 \\
  & \le \ltwo{\bbeta}^4 \cdot
    \sum_{r\le m/2} 16m^{-1}a(\bw_{0,r})^2 = 8\ltwo{\bbeta}^4 \cdot
    \frac{1}{m/2}\sum_{r\le m/2} a(\bw_{0,r})^2 \\
  & \le 32\pi[(p-2)^3\vee
    1]\alpha^2\Bx^{2(p-2)}\ltwo{\bbeta}^{2p}\delta^{-1}.
\end{align*}

\paragraph{Concentration of function}
Let $f_m(\bx) = \frac{1}{m}\sum_{r\le m/2}
\sigma''(\bw_{0,r}^\top\bx)a(\bw_{0,r})$. We now show the
concentration of $f_m$ to
$f_{\star,p-2}(\bx)\defeq \alpha(\beta^\top\bx)^{p-2}$ 
over the dataset
$\set{\bx_1,\dots,\bx_n}$. We perform a truncation
argument: let $R$ be a large radius (to be chosen) satisfying
\begin{equation}
  \P_{\bW_0}\paren{\sup_{r\in[m]} \ltwo{\bw_{0,r}}\ge R\Bx^{-1}} \ge 1 -
  \delta/2.
\end{equation}
On this event we have
\begin{equation*}
  f_m(\bx)=\frac{1}{m}\sum_{r\le
  m}\sigma''(\bw_{0,r}^\top\bx)a(\bw_{0,r})\indic{\ltwo{\bw_{0,r}}\le
  R\Bx^{-1}} \defeq f_m^R(\bx).
\end{equation*}
Letting $f_{\star,p-2}^R(\bx)\defeq
\E_{\bw_0}[\sigma''(\bw_{0}^\top\bx)a(\bw_0)\indic{\ltwo{\bw_0}\le
  R\Bx^{-1}}]$, we have
\begin{align*}
  \E_{\bW_0}\brac{ \paren{f_m(\bx) - f_{\star, p-2}^R(\bx)}^2} =
  \frac{1}{m}\E_{\bw_0}\brac{\sigma''(\bw_0^\top\bx)a^2(\bw_0)
  \indic{\ltwo{\bw_0}\le R}} \le C\frac{R^2R_a^2}{m}.
\end{align*}
Applying Chebyshev inequality and a union bound, we get
\begin{align*}
  \P\paren{\max_i \abs{f_m(\bx_i) - f_{\star, p-2}(\bx_i)} \ge t} \le
  C\frac{nR^2R_a^2}{mt^2}.
\end{align*}
For any $\eps>0$, by substituting in
$t=\eps\Bx^{-2}\ltwo{\bbeta}^{-2}/2$, we see that
\begin{equation}
  \label{equation:req-m}
  m \ge O\paren{nR^2R_a^2\Bx^4\ltwo{\bbeta}^4\eps^{-2}} =
  O\paren{nR^2(p-2)^3\alpha^2\Bx^{2p}\ltwo{\bbeta}^{2p}\eps^{-2}}
\end{equation}
ensures that
\begin{equation}
  \label{equation:f-sampling-bound}
  \max_{i\in[n]} |f_m(\bx_i) - f^R_{\star,p-2}(\bx_i)|\le
  \eps\Bx^{-2}\norm{\bbeta}^{-2}/2.
\end{equation}

Next, for any $\bx$ we have the bound
\begin{align*}
  & \quad \abs{f^R_{\star,p-2}(\bx) - f_{\star,p-2}(\bx)} =
    \abs{\E_{\bw_0}[\sigma''(\bw_0^\top\bx)a(\bw_0)\indic{\ltwo{\bw_0} >
    R}]} \\
  & \le \E[a(\bw_0)^2]^{1/2} \cdot \E[\sigma''(\bw_0^\top\bx)^4]^{1/4}
    \cdot \P(\ltwo{\bw_0}> R)^{1/4} \\
  & \le R_a \cdot C/\sqrt{d} \cdot \P(\ltwo{\bw_0}>R)^{1/4}.
\end{align*}
Choosing $R$ such that
\begin{equation}
  \label{equation:req-R}
  \P(\ltwo{\bw_0} > R) \le
  c\frac{\sqrt{d}\eps^4}{R_a\Bx^8\ltwo{\bbeta}^8}
\end{equation}
ensures that
\begin{equation}
  \label{equation:f-truncation-bound}
  \max_{i} \abs{f_{\star, p-2}^R(\bx_i) - f_{\star, p-2}(\bx_i)} \le
  \frac{\eps\Bx^{-2}\ltwo{\bbeta}^{-2}}{2}.
\end{equation}
Combining~\eqref{equation:f-sampling-bound}
and~\eqref{equation:f-truncation-bound}, we see that with probability
at least $1-\delta$,
\begin{align*}
  & \quad \max_{i\in[n]} \abs{f^Q_{\bW_\star}(\bx_i) - f_\star(\bx_i)}
    = \max_{i\in[n]} \abs{f_m(\bx_i) - f_{\star,p-2}(\bx_i)} \cdot
    (\bbeta^\top\bx_i)^2 \\
  & \le 2 \cdot \frac{\eps\Bx^{-2}\ltwo{\bbeta}^{-2}}{2} \cdot
    \Bx^2\ltwo{\bbeta}^2 = \eps
\end{align*}
and thus
\begin{align}
  \label{equation:existential-bound}
  \abs{L^Q(\bW_\star) - L(f_\star)} \le \eps.
\end{align}
To satisfy the requirements for $m$ and $R$ in~\eqref{equation:req-R}
and~\eqref{equation:req-m}, we first set $R=\wt{O}(\sqrt{d})$ (with
sufficiently large log factor) to satisfy~\eqref{equation:req-R} by
standard Gaussian norm concentration
(cf. Appendix~\ref{appendix:w0-bound}), and by~\eqref{equation:req-m}
it suffices to set $m$ as
\begin{equation*}
  m \ge \wt{O}\paren{nd(p-2)^3\alpha^2(\Bx\ltwo{\bbeta})^{2p}\eps^{-2}}.
\end{equation*}
for~\eqref{equation:existential-bound} to hold.
\end{proof-of-theorem}


\subsubsection{Proof of main theorem}
\label{appendix:expressivity-multiple}
We apply
Theorem~\ref{theorem:quad-net-expressivity} $k$ times: let
\begin{equation*}
  f_{\star,j}(\bx) \defeq \alpha_j(\bbeta_j^\top\bx)^{p_j},
\end{equation*}
so that $f_\star=\sum_{j\le k}f_{\star, j}$. Associate each $j$ with an
\emph{independent} set of initialization
$(\ba_0^{(j)},\bW_0^{(j)})$. By
Theorem~\ref{theorem:quad-net-expressivity}, there exists
$\bW_{\star}^{(j)}\in\R^{d\times m_j}$, where
\begin{equation*}
  m_j =
  \wt{O}\paren{ndk^2p_j^3\alpha_j^2(\Bx\ltwo{\bbeta_j})^{2p_j}\eps^{-2}} 
\end{equation*}
such that with probability at least $1-\delta/k$ we have
\begin{equation*}
  \max_{i\in[n]} \abs{f^Q_{\bW_\star^{(j)}}(\bx_i) - f_\star(\bx_i)} \le
  \eps/k
\end{equation*}
and the norm bound
\begin{equation*}
  \normtf{\bW_\star^{(j)}}^4 \le
  O\paren{kp_j^3\alpha_j^2\Bx^{2(p_j-2)}\ltwo{\bbeta_j}^{2p_j}\delta^{-1}}.
\end{equation*}
(Note we have slightly abused notation, so that now
$\set{f^Q_{\bW_\star^{(j)}}}_{j\in[k]}$ use a disjoint set of initial
weights $(\ba_0^{(j)}, \bW_0^{(j)})$.) Concatenating all the
$(\bW_\star^{(j)}, \ba_0^{(j)}, \bW_0^{(j)})$ and applying a union
bound, we have the following: so long as the width
\begin{equation*}
  m \ge \sum_{j=1}^k m_j = 
  \wt{O}\paren{ndk^2\sum_{j=1}^kp_j^3
    \alpha_j^2(\Bx\ltwo{\bbeta_j})^{2p_j}\eps^{-2}},
\end{equation*}
with probability at least $1-\delta$ (over $\ba_0\in\R^m$ and
$\bW_0\in\R^{d\times m}$), there exists
$\bW_\star\in\R^{d\times m}$ such that
\begin{equation*}
  \max_{i\in[n]} \abs{f^Q_{\bW_\star}(\bx_i) - f_\star(\bx_i)} \le
  \eps,
\end{equation*}
which by the 1-Lipschitzness of the loss implies that
\begin{equation*}
  L^Q(\bW_\star) \le L(f_\star) + \eps = \eps_0 + \eps.
\end{equation*}
Further, as $\bW_\star$ is the concatenation of
$\set{\bW_\star^{(j)}}_{j\in[k]}$, we have the norm bound
\begin{equation*}
  \normtf{\bW_\star}^4 = \sum_{j=1}^k \normtf{\bW_\star^{(j)}}^4 =
  \wt{O}\paren{k\sum_{j=1}^k
    p_j^3\alpha_j^2\Bx^{2(p_j-2)}\ltwo{\bbeta_j}^{2p_j}\delta^{-1}}.
\end{equation*}
This is the desired result.
\qed

\section{Existence, generalization, and expressivity of higher-order
  NTKs}
\label{appendix:higher-ntk-gen}
In this section we formally study the generalization and expressivity
of higher-order NTKs outlined in
Section~\ref{section:higher-order}. Let $k\ge 2$ be an integer, and
recall for any $\bW\in\R^{d\times 2m}$ the definition of the $k$-th
order NTK
\begin{equation}
  f^{(k)}_{\bW_0,\bW}(\bx) = \frac{1}{\sqrt{m}}\sum_{r\le m}
  \frac{\sigma^{(k)}(\bw_{0,r}^\top\bx)}{k!} \brac{(\bw_{+,r}^\top\bx)^k -
    (\bw_{-,r}^\top\bx)^k},
\end{equation}

\subsection{Coupling $f$ and $f^{(k)}$ via randomziation}
\label{appendix:higher-ntk-coupling}
Recall that for analytic $\sigma$ we have the expansion
\begin{equation*}
  f_{\bW_0+\bW}(\bx) = \frac{1}{\sqrt{m}}\sum_{r\le m}
  \sigma((\bw_{0,r}+\bw_{+,r})^\top\bx) -
  \sigma((\bw_{0,r}+\bw_{-,r})^\top\bx) =
  \sum_{k=0}^\infty f^{(k)}_{\bW_0,\bW}(\bx),
\end{equation*}
For an arbitrary $\bW$ such that
$\ltwo{\bw_{+,r}},\ltwo{\bw_{-,r}}=o_m(1)$, we expect that
$f^{(1)}(\bx)$ is the dominating term in the expansion.

\paragraph{Extracting the $k$-th order term}
We now describe an approach to finding $\bW$ so that
\begin{equation*}
  f_{\bW_0+\bW}(\bx) = f^{(k)}_{\bW_0, \bW}(\bx) + o_m(1),
\end{equation*}
that is, the neural net is approximately the $k$-th order NTK plus an
error term that goes to zero as $m\to\infty$, thereby ``escaping'' the
NTK regime. Our approach builds on the following randomization
technique: let $z_+$, $z_-$ be two random variables (distributions)
such that
\begin{equation*}
  \E[z_+^j] = \E[z_-^j]~~{\rm for}~j=0,1,\dots,k-1~~~{\rm
    and}~~~\E[z_+^k] = \E[z_-^k] + 1.
\end{equation*}
Set $(\bw_{+,r},\bw_{-,r}) = (z_{+,r}\bw_{\star,r},
z_{-,r}\bw_{\star,r})$, and take $\ltwo{\bw_{\star,r}}=O(m^{-1/2k})$, we have
\begin{align*}
  f^{(j)}_{\bW_0,\bW}(\bx) = \frac{1}{\sqrt{m}} \sum_{r\le
  m}\frac{1}{j!}\sigma^{(j)}(\bw_{0,r}^\top\bx) \underbrace{(z_{+,r}^j -
  z_{-,r}^j)}_{\textrm{mean
  zero}}\underbrace{(\bw_{\star,r}^\top\bx)^j}_{O(m^{-j/2k})} = O_p(m^{-j/2k})
\end{align*}
for all $j=1,\dots,k-1$, and
\begin{align*}
  & \quad f^{(k)}_{\bW_0,\bW}(\bx) = \frac{1}{\sqrt{m}} \sum_{r\le
    m}\frac{1}{k!}\sigma^{(k)}(\bw_{0,r}^\top\bx) \underbrace{(z_{+,r}^k -
    z_{-,r}^k)}_{\textrm{mean}=1}\underbrace{(\bw_{\star,r}^\top\bx)^k}_{O(m^{-1/2})}
      = O_P(1),
\end{align*}
and
\begin{align*}
  f^{(k+1)}_{\bW_0,\bW}(\bx) = \frac{1}{\sqrt{m}} \sum_{r\le
  m}\frac{1}{(k+1)!}\sigma^{(k+1)}(\bw_{0,r}^\top\bx) \underbrace{(z_{+,r}^{k+1} -
  z_{-,r}^{k+1})}\underbrace{(\bw_{\star,r}^\top\bx)^{k+1}}_{O(m^{-(k+1)/2k})}
  = O_P(m^{-1/2k}).
\end{align*}
Therefore, with high probability, all $f^{(1)},\dots,f^{(k-1)}$ as
well as the remainder term $f-\sum_{j\le k}f^{(j)}$ has order
$O(m^{-1/2k})$, and the $k$-th order NTK $f^{(k)}$ can express an
$O(1)$ function.

\subsection{Generalization and expressivity of $f^{(k)}$}
\label{appendix:higher-ntk-gen-expressivity}
We now turn to studying the generalization and expressivity of the
$k$-th order NTK $f^{(k)}$, Throughout this subsection, we assume
(for convenience) that
\begin{equation*}
  \sigma_k(t) \defeq \frac{1}{k!}\sigma^{(k)}(t) \equiv \relu(t)
\end{equation*}
is the ReLU activation. 

As we have seen in Section~\ref{section:higher-order}, we have
$f^{(k)}=O(1)$ by choosing $\bw_r\sim O(m^{-1/2k})$, therefore we
restrict attention on such $\bW$'s by considering the constraint set
$\{\bW:\normttk{\bW}^{2k}\le\Bw^{2k}\}$ for some $\Bw=O_m(1)$.

\paragraph{Overview of results}
This subsection establishes the following results for the $k$-th order
NTK. 
\begin{itemize}[wide, labelwidth=!, labelindent=0pt]
\item We bound the generalization of $f^{(k)}$ through the tensor
  operator norm of a certain $k$-tensor involving the features
  (Lemma~\ref{lemma:gen-tensornorm}).
  Consequently, the generalization of the $k$-th order NTK for
  $\normttk{\bW}\le \Bw$, when the base distribution of $\bx$ is
  uniform on the sphere, scales as
  \begin{align*}
    \wt{O}\paren{\Bx^k\Bw^k\brac{\frac{1}{\sqrt{nd^{k-1}}} +
    \frac{1}{n}} + \frac{1}{\sqrt{n}}}.
  \end{align*}
  (Theorem~\ref{theorem:gen-unif}).
  Compared with the distribution-free bound $\Bx^k\Bw^k/\sqrt{n}$, the
  leading term is better by a factor of $\sqrt{\min\set{d^{k-1}, n}}$. In
  particular, when $n\ge d^{k-1}$, the generalization is better by a
  factor of $\sqrt{d^{k-1}}$ than the distribution-free bound.
\item For the polynomial $f_\star(\bx)=\alpha(\bbeta^\top\bx)^p$ with
  $p\ge k$ (and $p-k$ is even or one), when $m$ is sufficiently large,
  there exists a $\bW_\star$ expressing $f_\star$ such that
  \begin{align*}
    \normttk{\bW_\star}^{2k} \le
    O\paren{p^3\alpha^2\Bx^{2(p-k)}\ltwo{\bbeta}^{2p}}.
  \end{align*}
  (Theorem~\ref{theorem:higher-ntk-expressivity}).  Substituting into
  the generalization bound yields the following generalization error
  for learning $f_\star$:
  \begin{align*}
    \wt{O}\paren{ p^3\alpha^2(\Bx\ltwo{\bbeta})^p \brac{
    \frac{1}{\sqrt{nd^{k-1}}} + \frac{1}{n}} }.
  \end{align*}
  In particular, the leading multiplicative factor is the same for all
  $k$ (including the linear NTK with $k=1$), but the sample complexity
  is lower by a factor of $d^{k-1}$ when $n\ge d^{k-1}$. This shows
  systematically the benefit of higher-order NTKs when distributional
  assumptions are present.
\end{itemize}

\paragraph{Tensor operator and nuclear norm}
Our result requires the definition of operator norm and
nuclear norm for $k$-tensors, which we briefly review here.
The operator norm of a symmetric $k$-tensor $\bA\in\R^{d^k}$ is
defined as
\begin{equation*}
  \opnorm{\bA} \defeq \sup_{\ltwo{\bv}=1} \<\bA, \bv^{\otimes k}\> =
  \sup_{\ltwo{\bv}=1} \bA[\bv,\dots,\bv].
\end{equation*}
The nuclear norm $\lnuc{\cdot}$ is defined as the dual norm of the
operator norm:
\begin{equation*}
  \lnuc{\bA} \defeq \sup_{\opnorm{\bB}=1} \<\bA, \bB\>.
\end{equation*}
Specifically, for any rank-one tensor $\bu^{\otimes k}$, we have
\begin{equation*}
  \lnuc{\bu^{\otimes k}} = \sup_{\opnorm{\bB}=1} \<\bu^{\otimes k},
  \bB\> = \ltwo{\bu}^k,
\end{equation*}
i.e. its nuclear norm equals its operator norm (and also the
Frobenius norm).

\subsubsection{Generalization}
We begin by stating a generalization bound for $f^{(k)}$, which depends
on the operator norm of a $k$-th order tensor feature, generalizing
Lemma~\ref{lemma:gen-opnorm}.
\begin{lemma}[Bounding generalization of $f^{(k)}$ via tensor
  operator norm]
  \label{lemma:gen-tensornorm}
  For any non-negative loss $\ell$ such that $z\mapsto \ell(y, z)$ is
  1-Lipschitz and $\ell(y,0)\le 1$ for all $y\in\mc{Y}$, we have the
  Rademacher complexity bound
  \begin{align*}
        \E_{\bsigma, \bx}\brac{\sup_{\normttk{\bW}\le \Bw}
        \frac{1}{n}\sum_{i=1}^n \sigma_i\ell(y_i, f^{(k)}_{\bW_0,\bW}(\bx_i))}
        \le 2\Bw^k\E_{\bsigma,\bx}\brac{\max_{r\in[m]}
        \opnorm{\frac{1}{n}\sum_{i=1}^n\sigma_i\sigma_k(\bw_{0,r}^\top\bx_i)
        \bx_i^{\otimes k}}} + \frac{1}{\sqrt{n}},
  \end{align*}
  where $\sigma_i\simiid\Unif\set{\pm 1}$ are Rademacher variables.
\end{lemma}
\begin{proof}
  The proof is analogous to that of Lemma~\ref{lemma:gen-opnorm}.
  As the loss $\ell(y,z)$ is 1-Lipschitz in $z$ for all $y$, by the
  Rademacher contraction theorem~\citep[Chapter 5]{wainwright2019high}
  we have that
  \begin{align*}
    & \quad \E_{\bsigma, \bx}\brac{\sup_{\normttk{\bW}\le \Bw}
      \frac{1}{n}\sum_{i=1}^n \sigma_i\ell(y_i,
      f^{(k)}_{\bW_0,\bW}(\bx_i))} \\
    & \le 2\E_{\bsigma,
      \bx}\brac{\sup_{\normttk{\bW}\le \Bw} \frac{1}{n}\sum_{i=1}^n
      \sigma_i f^{(k)}_{\bW_0,\bW}(\bx_i)} + \E_{\bsigma,
      \bx}\brac{\frac{1}{n}\sum_{i=1}^n
      \sigma_i\ell(y_i, 0)} \\
    & \le 2\E_{\bsigma,\bx}\brac{\sup_{\normttk{\bW}\le \Bw}
      \frac{1}{\sqrt{m}}\sum_{r\le m} 
      \<\frac{1}{n}\sum_{i=1}^n\sigma_i
      a_r\sigma_k(\bw_{0,r}^\top\bx_i)\bx_i^{\otimes k},
      \bw_r^{\otimes k}\>} + \frac{1}{\sqrt{n}} \\
    & \le 2\E_{\bsigma,
      \bx}\brac{\sup_{\normttk{\bW}\le \Bw}
      \max_{r\in[m]}\opnorm{\frac{1}{n}\sum_{i=1}^n
      a_r\sigma_i\sigma_k(\bw_{0,r}^\top\bx_i)\bx_i^{\otimes k}} \cdot
      \frac{1}{\sqrt{m}}\sum_{r\le m}\lnuc{\bw_r^{\otimes k}}} +
      \frac{1}{\sqrt{n}}\\ 
    & \le 2\E_{\bsigma,\bx}\brac{\max_{r\in[m]}\opnorm{\frac{1}{n}\sum_{i=1}^n
      \sigma_i\sigma_k(\bw_{0,r}^\top\bx_i)\bx_i^{\otimes k}}} \cdot
      \underbrace{\sup_{\normttk{\bW}\le \Bw}
      \frac{1}{\sqrt{m}}\ltwo{\bw_r}^{k}}_{\le \Bw^k} +
      \frac{1}{\sqrt{n}},
  \end{align*}
  where the last step used the power mean (or Cauchy-Schwarz)
  inequality on $\set{\ltwo{\bw_r}}$. \qed
\end{proof}

\paragraph{Bound on tensor operator norm}
It is straightforward to see that the expected tensor operator norm can
be bounded as
\begin{equation*}
  \wt{O}\paren{\Bx^k/\sqrt{n}}
\end{equation*}
without any distributional assumptions on $\bx$.
We now provide a bound on the expected tensor operator norm appearing
in Lemma~\ref{lemma:gen-tensornorm} in the special case of uniform
features, i.e. $\bx\sim\Unif(\S^{d-1}(\Bx))$.

\begin{lemma}[Tensor operator norm bound for uniform features]
  \label{lemma:unif-tensornorm-bound}
  Suppose $\bx_i\simiid \Unif(\S^{d-1}(\Bx))$. Then for any $k\ge 3$
  and $k=O(1)$, we have
  (with high probability over $\bW_0$)
  \begin{equation}
    \E_{\bsigma, \bx}\brac{\max_{r\in[m]}
      \opnorm{\frac{1}{n}\sum_{i=1}^n
        \sigma_i\sigma_k(\bw_{0,r}^\top\bx_i)\bx_i^{\otimes k}}} \le
    \wt{O}\paren{\Bx^k\brac{\frac{1}{\sqrt{nd^{k-1}}} +
        \frac{1}{n}}}.
  \end{equation}
\end{lemma}
Substituting the above bound into Lemma~\ref{lemma:gen-tensornorm}
directly leads to the following generalization bound for $f^{(k)}$:
\begin{theorem}[Generalization for $f^{(k)}$ with uniform features]
  \label{theorem:gen-unif}
  Suppose $\bx_i\simiid \Unif(\S^{d-1}(\Bx))$. Then for any $k\ge 3$
  and $k=O(1)$, we have (with high
  probability over $\bW_0$)
  \begin{align*}
    \E_\cD\brac{ \sup_{\normttk{\bW}\le \Bw}
    \paren{L_P(f^{(k)}_{\bW_0,\bW}) - L(f^{(k)}_{\bW_0,\bW})}} \le
    \wt{O}\paren{ \Bx^k\Bw^k\brac{ \frac{1}{\sqrt{nd^{k-1}}} +
    \frac{1}{n}} + \frac{1}{\sqrt{n}}}.
  \end{align*}
\end{theorem}
The proof of Lemma~\ref{lemma:unif-tensornorm-bound} is deferred to
Appendix~\ref{appendix:proof-unif-tensornorm-bound}.

\subsubsection{Expressivity}
\begin{theorem}[Expressivity of $f^{(k)}$]
  \label{theorem:higher-ntk-expressivity}
  Suppose $\set{(a_r, \bw_{0,r})}$ are generated according to the
  symmetric initialization~\eqref{equation:symmetric-init}, and
  $f_\star(\bx)=\alpha(\bbeta^\top\bx)^p$ where
  $p-k\in\set{1}\cup\set{2\ell}_{\ell\ge 0}$. Suppose further that
  $\sigma$ is such that $\sigma_k(t)=\relu(t)$, then so long as the
  width is sufficiently large:
  \begin{equation*}
    m\ge
    \wt{O}\paren{ndp^3\alpha^2(\Bx\ltwo{\bbeta})^{2p}\eps^{-2}},
  \end{equation*}
  we have with probability at least $1-\delta$ (over $\bW_0$) that
  there exists $\bW_\star\in\R^{d\times m}$ such that 
  \begin{equation*}
    \abs{L^Q(\bW_\star) - L(f_\star)} \le \eps~~~{\rm and}~~~
    \normttk{\bW_\star}^{2k} \le \Bwstar^{2k} =
    O\paren{p^3\alpha^2\Bx^{2(p-k)}\ltwo{\bbeta}^{2p}\delta^{-1}}. 
  \end{equation*}
\end{theorem}
The proof of Theorem~\ref{theorem:higher-ntk-expressivity} is deferred
to Appendix~\ref{appendix:proof-higher-ntk-expressivity}.

\subsection{Proof of Lemma~\ref{lemma:unif-tensornorm-bound}}
\label{appendix:proof-unif-tensornorm-bound}
We begin by observing for any symmetric tensor $\bA\in\R^{d^k}$ that
\begin{equation*}
  \opnorm{\bA} \le \frac{1}{1-k\eps}\sup_{\bv\in N(\eps)}
  \bA[\bv,\dots,\bv],
\end{equation*}
where $N(\eps)$ is an $\eps$-covering of unit sphere $\S^{d-1}(1)$.
(The proof follows by bounding $\bA[\bu,\dots,\bu]$ by
$\bA[\bv,\dots,\bv] + k\eps\opnorm{\bA}$ through replacing $\bu$ by
$\bv$ one at a time). Taking $\eps=1/(2k)$, we have
\begin{align*}
  & \quad \P_{\bsigma,\bx}\paren{\max_{r\in[m]} \opnorm{\frac{1}{n}\sum_{i=1}^n
    \sigma_i\sigma_k(\bw_{0,r}^\top\bx_i)\bx_i^{\otimes k}} \ge t} \\
  & \le \P_{\bsigma,\bx}\paren{\max_{r\in[m], \bv\in N(1/(2k))}
    \frac{1}{n}\sum_{i=1}^n
    \sigma_i\sigma_k(\bw_{0,r}^\top\bx_i)(\bv^\top\bx_i)^k \ge t/2}.
\end{align*}
We now perform a truncation argument to upper bound the above
probability. Let $M>0$ be a truncation level to be determined, we have
by the Bernstein inequality that
\begin{align*}
  & \quad \P_{\bsigma,\bx}\paren{\max_{r\in[m]} \opnorm{\frac{1}{n}\sum_{i=1}^n
    \sigma_i\sigma_k(\bw_{0,r}^\top\bx_i)\bx_i^{\otimes k}} \ge t} \\
  & \le \P_{\bsigma,\bx}\paren{\max_{r\in[m], \bv\in N(1/(2k))}
    \frac{1}{n}\sum_{i=1}^n
    \sigma_i\sigma_k(\bw_{0,r}^\top\bx_i)(\bv^\top\bx_i)^k
    \indic{\abs{\sigma_k(\bw_{0,r}^\top\bx_i)}\le
    M} \ge t/2} \\
  & \qquad + \P_{\bx} \paren{ \max_{r,i}
    \abs{\sigma_k(\bw_{0,r}^\top\bx_i)} \ge M } \\
  & \le \exp\paren{-c \min\set{\frac{nt^2}{\wt{O}(1)\cdot
    \Bx^{2k}d^{-k}}, 
    \frac{nt}{M\Bx^k}} + d\log 6k + \log m} +
    \exp\paren{-\frac{M^2}{2\wt{O}(1)} + \log mn},
\end{align*}
where the $\wt{O}(1)\cdot \Bx^{2k}d^{-k}$ comes from computing the
variance of
\begin{equation*}
  Z_i \defeq \sigma_i\sigma_k(\bw_{0,r}^\top\bx_i)(\bv^\top\bx_i)^k
\end{equation*}
using that $\bx_i$ are uniform on the sphere (see,
e.g.~\citep[Proof of Lemma 4]{ghorbani2019linearized}); $MB_x^k$ is
the bound on the variable $Z_i$, and the $\wt{O}(1)$
comes from the fact that $\ltwo{\bw_{0,r}}\le 
\wt{O}(\sqrt{d}\Bx^{-1})$ with high probability. Now, choosing 
\begin{equation*}
  M = (nt/\Bx^k)^{1/2},
\end{equation*}
the above bound reads
\begin{align*}
  \exp\paren{-c \min\set{\frac{nt^2}{\wt{O}(1)\cdot
    \Bx^{2k}d^{-k}}, 
    \paren{\frac{nt}{\Bx^k}}^{1/2}} + \wt{O}(d)} +
    \exp\paren{-\frac{nt/\Bx^k}{2\wt{O}(1)} + \wt{O}(1)}
  \defeq p_t.
\end{align*}
It remains to bound $\int_{t=0}^\infty p_t$ to give an expectation
bound on the desired tensor operator norm. This follows by adding up
the following three bounds:
\begin{enumerate}[(1)]
\item For the main branch ``$nt^2/\wt{O}(\Bx^{2k}d^{-k})$'' we have
  \begin{align*}
    \int_0^\infty
    \min\set{\exp\paren{-\frac{nt^2}{\wt{O}(\Bx^{2k}d^{-k})} + 
    \wt{O}(d)}, 1}dt \le \wt{O}\paren{\sqrt{\frac{\Bx^{2k}}{nd^{k-1}}}}.
  \end{align*}
  This follows by integrating the ``1'' branch for
  $t\le \wt{O}(\sqrt{\Bx^{2k}d^{-(k-1)}/n})$ (which yields the right
  hand side) and integrating the other branch otherwise (the integral
  being upper bounded by $\wt{O}(\sqrt{\Bx^{2k}d^{-k}/n})$, dominated
  by the right hand side).
\item The branch ``$(nt/\Bx^k)^{1/2}$" is taken only when
  \begin{equation*}
    \paren{ \frac{nt}{\Bx^k} }^{1/2} <
    \frac{nt^2}{\wt{O}(\Bx^{2k}d^{-k})}~~~{\rm i.e.}~~~t >
    \wt{O}\paren{n^{-1/3}\Bx^kd^{2k/3}}.
  \end{equation*}
  On the other hand, the inequality $(nt/\Bx^k)^{1/2}>\wt{O}(d)$
  happens when
  \begin{equation*}
    t > \wt{O}\paren{ d^2\Bx^k/n},
  \end{equation*}
  which is implied by the preceding condition so long as $k\ge
  3$. Therefore, when this branch is taken, the $\wt{O}(d)$ can
  already be absorbed into the main term, so the contribution of this
  branch can be bounded as
  \begin{align*}
    \int_{\wt{O}(n^{-1/3}\Bx^kd^{2k/3})}^\infty
    \exp\paren{-c'\paren{\frac{nt}{\Bx^k}}^{1/2}} dt
    \le \int_0^\infty
    \exp\paren{-c'\paren{\frac{nt}{\Bx^k}}^{1/2}} dt \le
    O\paren{\frac{\Bx^k}{n}}.
  \end{align*}
\item We have
  \begin{align*}
    \int_0^\infty
    \min\set{\exp\paren{-\frac{nt/\Bx^k}{2\wt{O}(1)} +
    \wt{O}(1)}, 1}dt \le \wt{O}\paren{\Bx^k/n},
  \end{align*}
  using a similar argument as part (1).
\end{enumerate}
Putting together the above three bounds, we obtain
\begin{align*}
  \max_{r\in[m]} \opnorm{\frac{1}{n}\sum_{i=1}^n
  \sigma_i\sigma_k(\bw_{0,r}^\top\bx_i)\bx_i^{\otimes k}} \le
  \int_0^\infty p_tdt \le \wt{O}\paren{ \Bx^k\brac{
  \frac{1}{\sqrt{nd^{k-1}}} + \frac{1}{n} }},
\end{align*}
the desired result. \qed

\subsection{Proof of Theorem~\ref{theorem:higher-ntk-expressivity}}
\label{appendix:proof-higher-ntk-expressivity}
Our proof is analogous to that of
Theorem~\ref{theorem:quad-net-expressivity}, in which we first look
at the case of infinitely many neurons and then use concentration to
carry the result onto finitely many neurons.

\paragraph{Expressivity with infinitely many neurons}
We first consider expressing $f_\star(\bx)=\alpha(\bbeta^\top\bx)^p$
with infinite-neuron version of $f^{(k)}$, that is, we wish to find
random variables $(\bw_+,\bw_-)$ such that
\begin{equation*}
  \E_{\bw_0}\brac{\relu(\bw_0^\top\bx) \paren{(\bw_+^\top\bx)^k -
      (\bw_-^\top\bx)^k}} = f_\star(x).
\end{equation*}
Choosing
\begin{equation*}
  (\bw_+, \bw_-) = \paren{([a]_+)^{1/k}\bbeta, ([a]_-)^{1/k}\bbeta}
\end{equation*}
for some real-valued random scalar $a$ (that depends on $\bw_0$),
we have
\begin{equation*}
  \E_{\bw_0}\brac{\relu(\bw_0^\top\bx) \paren{(\bw_+^\top\bx)^k -
      (\bw_-^\top\bx)^k}} = (\bbeta^\top\bx)^k \cdot
  \E_{\bw_0}\brac{\relu(\bw_0^\top\bx) a},
\end{equation*}
therefore the task reduces to finding $a=a(\bw_0)$ such that
$\E_{\bw_0}\brac{\relu(\bw_0^\top\bx)
  a}=\alpha(\bbeta^\top\bx)^{p-k}$. By
Lemma~\ref{lemma:relu-rf-expressivity}, there exists $a=a(\bw_0)$
satisfying the above and such that
\begin{equation}
  \label{equation:higher-ntk-a-bound}
  \E_{\bw_0}[a^2] \le 2\pi((p-k)\vee
  1)^3\alpha^2\Bx^{2(p-k)}\ltwo{\bbeta}^{2(p-k)}.
\end{equation}
Using this $a$, the $k$-th order NTK defined by $(\bw_+,\bw_-)$
expresses $f_\star$ and further satisfies the bound
\begin{align*}
  \E_{\bw_0}\brac{\ltwo{\bw_+}^{2k} + \ltwo{\bw_-}^{2k}} =
  \E{\bw_0}[a^2] \cdot \ltwo{\bbeta}^{2k} \le 2\pi((p-k)\vee
  1)^3\alpha^2\Bx^{2(p-k)}\ltwo{\bbeta}^{2p}.
\end{align*}

\paragraph{Finite neurons}
Given the symmetric initialization $\set{\bw_{0,r}}_{r=1}^m$, for all
$r\in[m]$, we consider $\bW_\star\in\R^{d\times m}$ defined through
\begin{equation*}
  (\bw_{\star,r}, \bw_{\star,r+m}) =
  \paren{m^{-1/2k}\bw_+(\bw_{0,r}),
    m^{-1/2k}\bw_-(\bw_{0,r})},
\end{equation*}
where we recall $(\bw_+(\bw_0), \bw_-(\bw_0)) =
(a_+(\bw_0)^{1/k}\bbeta, a_-(\bw_0)^{1/k}\bbeta)$.
We then have
\begin{align*}
  & \quad f^{(k)}_{\bW_0,\bW_\star}(\bx) = \frac{1}{\sqrt{m}}\sum_{r\le
    m}\sigma_k(\bw_{0,r}^\top\bx) \brac{(\bw_{\star,r}^\top\bx)^k -
    (\bw_{\star,r+m}^\top\bx)^k} \\
  & = \frac{1}{m}\sum_{r\le m} \sigma_k(\bw_{0,r}^\top\bx) \brac{
    (\bw_+(\bw_{0,r})^\top\bx)^k - (\bw_-(\bw_{0,r})^\top\bx)^k } \\
  & = \brac{\frac{1}{m}\sum_{r\le m} \sigma_k(\bw_{0,r}^\top\bx)
    a(\bw_{0,r})} \cdot (\bbeta^\top\bx)^k.
\end{align*}

\paragraph{Bound on $\normttk{\bW_\star}$}
As
$f_\star(\bx)=\alpha(\bbeta^\top\bx)^p$,~\eqref{equation:higher-ntk-a-bound}
guarantees that the coefficient $a(\bw_0)$ involved above satisfies
that
\begin{equation*}
  R_a^2 \defeq \E_{\bw_0}[a(\bw_0)^2] \le 2\pi((p-k)\vee
  1)^3\alpha^2\Bx^{2(p-k)}\ltwo{\bbeta}^{2(p-k)}.
\end{equation*}
By Markov inequality, we have with probability at least $1-\delta/2$
that
\begin{equation*}
  \frac{1}{m}\sum_{r\le m} a(\bw_{0,r})^2 \le 4\pi((p-k)\vee
  1)^3\alpha^2\Bx^{2(p-k)}\ltwo{\bbeta}^{2(p-k)}\delta^{-1},
\end{equation*}
which yields the bound
\begin{align*}
  & \quad \normttk{\bW}^{wk} = \sum_{r\le 2m}\ltwo{\bw_{\star, r}}^{2k} \\
  & \le \ltwo{\bbeta}^{2k} \cdot
    \sum_{r\le m} m^{-1}a(\bw_{0,r})^2 
  \le 4\pi[(p-k)^3\vee
  1]\alpha^2\Bx^{2(p-k)}\ltwo{\bbeta}^{2p}\delta^{-1}.
\end{align*}

\paragraph{Concentration of function}
Let $f_m(\bx) = \frac{1}{m}\sum_{r\le m}
\sigma_k(\bw_{0,r}^\top\bx)a(\bw_{0,r})$. We now show the
concentration of $f_m$ to
$f_{\star,p-k}(\bx)\defeq \alpha(\beta^\top\bx)^{p-k}$ 
over the dataset
$\set{\bx_1,\dots,\bx_n}$. We perform a truncation
argument: let $R$ be a large radius (to be chosen) satisfying
\begin{equation}
  \P_{\bW_0}\paren{\sup_{r\in[m]} \ltwo{\bw_{0,r}}\ge R\Bx^{-1}} \ge 1 -
  \delta/2.
\end{equation}
On this event we have
\begin{equation*}
  f_m(\bx)=\frac{1}{m}\sum_{r\le
    m}\sigma_k(\bw_{0,r}^\top\bx)a(\bw_{0,r})\indic{\ltwo{\bw_{0,r}}\le
    R\Bx^{-1}} \defeq f_m^R(\bx).
\end{equation*}
Letting $f_{\star,p-k}^R(\bx)\defeq
\E_{\bw_0}[\sigma_k(\bw_{0}^\top\bx)a(\bw_0)\indic{\ltwo{\bw_0}\le
  R\Bx^{-1}}]$, we have
\begin{align*}
  \E_{\bW_0}\brac{ \paren{f_m(\bx) - f_{\star, p-k}^R(\bx)}^2} =
  \frac{1}{m}\E_{\bw_0}\brac{\sigma_k(\bw_0^\top\bx)a^2(\bw_0)
  \indic{\ltwo{\bw_0}\le R}} \le C\frac{R^2R_a^2}{m}.
\end{align*}
Applying Chebyshev inequality and a union bound, we get
\begin{align*}
  \P\paren{\max_i \abs{f_m(\bx_i) - f_{\star, p-k}(\bx_i)} \ge t} \le
  C\frac{nR^2R_a^2}{mt^2}.
\end{align*}
For any $\eps>0$, by substituting in
$t=\eps\Bx^{-k}\ltwo{\bbeta}^{-k}/2$, we see that
\begin{equation}
  \label{equation:req-m}
  m \ge O\paren{nR^2R_a^2\Bx^{2k}\ltwo{\bbeta}^{2k}\eps^{-2}} =
  O\paren{nR^2(p-k)^3\alpha^2\Bx^{2p}\ltwo{\bbeta}^{2p}\eps^{-2}}
\end{equation}
ensures that
\begin{equation}
  \label{equation:f-sampling-bound}
  \max_{i\in[n]} |f_m(\bx_i) - f^R_{\star,p-2}(\bx_i)|\le
  \eps\Bx^{-k}\norm{\bbeta}^{-k}/2.
\end{equation}

Next, for any $\bx$ we have the bound
\begin{align*}
  & \quad \abs{f^R_{\star,p-k}(\bx) - f_{\star,p-k}(\bx)} =
    \abs{\E_{\bw_0}[\sigma_k(\bw_0^\top\bx)a(\bw_0)\indic{\ltwo{\bw_0} >
    R}]} \\
  & \le \E[a(\bw_0)^2]^{1/2} \cdot \E[\sigma_k(\bw_0^\top\bx)^4]^{1/4}
    \cdot \P(\ltwo{\bw_0}> R)^{1/4} \\
  & \le R_a \cdot C/\sqrt{d} \cdot \P(\ltwo{\bw_0}>R)^{1/4}.
\end{align*}
Choosing $R$ such that
\begin{equation}
  \label{equation:req-R}
  \P(\ltwo{\bw_0} > R) \le
  c\frac{\sqrt{d}\eps^4}{R_a\Bx^{4k}\ltwo{\bbeta}^{4k}}
\end{equation}
ensures that
\begin{equation}
  \label{equation:f-truncation-bound}
  \max_{i} \abs{f_{\star, p-2}^R(\bx_i) - f_{\star, p-2}(\bx_i)} \le
  \frac{\eps\Bx^{-k}\ltwo{\bbeta}^{-k}}{2}.
\end{equation}
Combining~\eqref{equation:f-sampling-bound}
and~\eqref{equation:f-truncation-bound}, we see that with probability
at least $1-\delta$,
\begin{align*}
  & \quad \max_{i\in[n]} \abs{f^Q_{\bW_\star}(\bx_i) - f_\star(\bx_i)}
    = \max_{i\in[n]} \abs{f_m(\bx_i) - f_{\star,p-k}(\bx_i)} \cdot
    (\bbeta^\top\bx_i)^k \\
  & \le 2 \cdot \frac{\eps\Bx^{-k}\ltwo{\bbeta}^{-k}}{2} \cdot
    \Bx^k\ltwo{\bbeta}^k = \eps
\end{align*}
and thus
\begin{align}
  \label{equation:existential-bound}
  \abs{L^Q(\bW_\star) - L(f_\star)} \le \eps.
\end{align}
To satisfy the requirements for $m$ and $R$ in~\eqref{equation:req-R}
and~\eqref{equation:req-m}, we first set $R=\wt{O}(\sqrt{d})$ (with
sufficiently large log factor) to satisfy~\eqref{equation:req-R} by
standard Gaussian norm concentration
(cf. Appendix~\ref{appendix:w0-bound}), and by~\eqref{equation:req-m}
it suffices to set $m$ as
\begin{equation*}
  m \ge \wt{O}\paren{nd[(p-k)^3\vee
    1]\alpha^2(\Bx\ltwo{\bbeta})^{2p}\eps^{-2}}.
\end{equation*}
for~\eqref{equation:existential-bound} to hold. \qed

\end{document}